\documentclass{article}
\pdfoutput=1

\usepackage{times}
\usepackage{graphicx} 
\usepackage{subfigure} 

\usepackage{natbib}

\usepackage{algorithm}
\usepackage{algorithmic}

\usepackage{hyperref}

\usepackage[accepted]{icml2017}

\icmltitlerunning{An Infinite HMM With Similarity-Biased Transitions}

\usepackage{amsmath}
\usepackage{amssymb}

\newcommand{\Norm}[2]{\mathcal{N}(#1,#2)}
\newcommand{\Unif}[1]{\mathsf{Unif}(#1)}
\newcommand{\Pois}[1]{\mathsf{Poisson}(#1)}
\newcommand{\Exp}[1]{\mathsf{Exp}(#1)}
\newcommand{\Gamm}[2]{\mathsf{Gamma}(#1,#2)}

\newcommand{\Be}[1]{\mathsf{Beta}\left(#1\right)}
\newcommand{\Binom}[2]{\mathcal{B}\mathrm{inom}(#1,#2)}

\newcommand{\Cat}[1]{\mathsf{Cat}(#1)}

\newcommand{\DP}[1]{\mathsf{DP}(#1)}
\newcommand{\GEM}[1]{\mathsf{GEM}(#1)}

\newcommand{\given}{\, \vert \,}

\newcommand{\abs}[1]{\left\vert #1 \right\vert}

\newcommand{\by}{\mathbf{y}}
\newcommand{\bY}{\mathbf{Y}}

\newcommand{\bM}{\mathbf{M}}

\newcommand{\bQ}{\mathbf{Q}}

\newcommand{\bz}{\mathbf{z}}

\newcommand{\bu}{\mathbf{u}}
\newcommand{\br}{\mathbf{r}}

\newcommand{\bw}{\mathbf{w}}
\newcommand{\bW}{\mathbf{W}}

\newcommand{\bbeta}{\boldsymbol{\beta}}
\newcommand{\btheta}{\boldsymbol{\theta}}
\newcommand{\bpi}{\boldsymbol{\pi}}
\newcommand{\bphi}{\boldsymbol{\phi}}

\newcommand{\bell}{\boldsymbol{\ell}}
\newcommand{\boldeta}{\boldsymbol{\eta}}
\newcommand{\bSigma}{\boldsymbol{\Sigma}}

\begin{document}
\twocolumn[

\icmltitle{ An Infinite Hidden Markov Model With Similarity-Biased
  Transitions}

\begin{icmlauthorlist}
\icmlauthor{ Colin Reimer Dawson }{oberlin}
\icmlauthor{ Chaofan Huang }{oberlin}
\icmlauthor{ Clayton T. Morrison }{ua}
\end{icmlauthorlist}

\icmlaffiliation{oberlin}{ Oberlin College, Oberlin, OH, USA}
\icmlaffiliation{ua}{The University of Arizona, Tucson, AZ, USA}
\icmlcorrespondingauthor{Colin Reimer Dawson}{cdawson@oberlin.edu}
\codelink{http://colindawson.net/hdp-hmm-lt}

\icmlkeywords{ Bayesian nonparametrics, Hidden Markov models }
\vskip 0.3in
]



\printAffiliationsAndNotice{}  

\begin{abstract}
We describe a generalization of the Hierarchical Dirichlet Process Hidden Markov Model (HDP-HMM) which is able to encode prior information that state transitions are more likely between ``nearby'' states.  This is accomplished by defining a similarity function on the state space and scaling transition probabilities by pairwise similarities, thereby inducing correlations among the transition distributions. 
We present an augmented data representation of the model as a Markov Jump Process in which: (1) some jump attempts fail, and (2) the probability of success is proportional to the similarity between the source and destination states.  This augmentation restores conditional conjugacy and admits a simple Gibbs sampler.
We evaluate the model and inference method on a speaker diarization task and a ``harmonic parsing'' task using four-part chorale data, as well as on several synthetic datasets, achieving favorable comparisons to existing models.

\end{abstract}

\section{Introduction and Background}

The 
hierarchical Dirichlet process hidden Markov model
(HDP-HMM) \cite{beal2001infinite, teh2006hierarchical} 
is a Bayesian model for time series data that generalizes 
the conventional hidden Markov
Model to allow a countably infinite state space.  
The hierarchical structure ensures that, despite the infinite state
space, a common set of destination states will be reachable with positive probability
from each source state.  The HDP-HMM can be characterized by the
following generative process.

Each state, indexed by $j$, has parameters,
$\theta_j$, drawn from a base measure, $H$.  A top-level
sequence of state weights, $\bbeta = (\beta_1, \beta_2, \dots)$, is
drawn by iteratively breaking a ``stick'' off of the remaining weight
according to a $\Be{1,\gamma}$ distribution.
The parameter $\gamma > 0$ is known as the concentration parameter and
governs how quickly the weights tend to decay, with 
large $\gamma$ corresponding to slow decay,
and hence more weights needed before a given cumulative weight is
reached.  This stick-breaking process 
is denoted by $\mathsf{GEM}$
\cite{ewens1990population,sethuraman1994constructive} 
for Griffiths, Engen and McCloskey.  
We thus have a discrete probability measure,
$G_0$, with weights $\beta_j$ at locations $\theta_j$, 
$j = 1, 2, \dots$, defined by
\begin{equation}
\theta_j \stackrel{\text{i.i.d.}}{\sim} H \qquad \bbeta \sim \GEM{\gamma}.
\end{equation}
$G_0$ drawn in this way is a Dirichlet Process (DP) random measure
with concentration $\gamma$ and base measure $H$.

The actual transition distribution, $\bpi_j$, from state $j$,
is drawn from another DP with concentration $\alpha$ and base measure $G_0$:
\begin{equation}
  \label{eq:1}
  \bpi_j \stackrel{\text{i.i.d.}}{\sim} \DP{\alpha G_0} \qquad j
  = 0, 1, 2, \dots
\end{equation}
where $\bpi_0$ represents the initial distribution. 
The hidden state sequence, $z_1, z_2, \dots z_T$ is then generated according to
$z_1 \given \bpi_0 \sim \Cat{\bpi_0}$, and 
\begin{equation}
  \label{eq:4}
  z_t \given z_{t-1}, \bpi_{z_{t-1}} \sim \Cat{\bpi_{z_{t-1}}} \qquad t = 1, 2, \dots, T
\end{equation}
Finally, the emission distribution 
for state $j$ is a function of $\theta_j$, so that observation
$y_t$ is drawn according to
\begin{equation}
  \label{eq:5}
  y_t \given z_{t}, \theta_{z_t} \sim F(\theta_{z_t})
\end{equation}

A shortcoming of the HDP prior on the transition matrix 
is that it does not use the fact that the source and destination
states are the same set: that is, each $\bpi_j$
has a special element which corresponds to a self-transition.  In the HDP-HMM, however,
self-transitions are no more likely {\it a priori} than
transitions to any other state.  The Sticky HDP-HMM \cite{fox2008hdp}
addresses this issue by adding an extra mass $\kappa$ at location $j$ to the base
measure of the DP that generates $\bpi_j$.  That is, \eqref{eq:1} is replaced by
\begin{equation}
  \label{eq:6}
  \bpi_j \sim \DP{\alpha G_0 + \kappa \delta_{\theta_j}}.
\end{equation}
An alternative approach that treats self-transitions as special 
is the HDP Hidden Semi-Markov Model (HDP-HSMM;
\citet{johnson2013bayesian}), wherein state duration distributions are modeled
separately, and ordinary self-transitions are ruled out.  
However, while both of these models have the ability to privilege
self-transitions, they contain no notion of
similarity for pairs of states that are not identical: 
in both cases, when the transition matrix
is integrated out, the prior probability of
transitioning to state $j'$ depends only on the top-level stick
weight associated with state $j'$, and not on the identity or
parameters of the previous state $j$.

The two main contributions of this paper are (1) a 
generalization of the HDP-HMM, which we call the HDP-HMM with local
transitions (HDP-HMM-LT) that allows for a
geometric structure to be defined on the latent state space, so that
``nearby'' states are {\em a priori} more likely to have transitions between
them, and (2) a simple Gibbs sampling algorithm for this model.  
The ``LT'' property is introduced by 
elementwise rescaling and then renormalizing of the HDP transition
matrix.  Two versions of the
similarity structure are illustrated: in one case, two states are
similar to the extent that their emission distributions are similar.
In another, the similarity structure is inferred separately.
In both cases, we give augmented data representations that restore
conditional conjugacy and thus allow a simple Gibbs sampling algorithm
to be used for inference.

A rescaling and renormalization approach similar to the one used in
the HDP-HMM-LT is used by \citet{paisley2012discrete} to define their
Discrete Infinite Logistic Normal (DILN) model, an instance of a 
correlated random measure \cite{ranganath2016correlated}, in the setting of
topic modeling.  There, however, the contexts and the mixture components
(topics) are distinct sets, and there is no notion of temporal
dependence. \citet{zhu2016hidden} developed an HMM
based directly on the DILN model\footnote{We thank an anonymous
  ICML reviewer for bringing this paper to our attention.}. Both
\citeauthor{paisley2012discrete} and \citeauthor{zhu2016hidden} employ variational
approximations, whereas we present a Gibbs sampler, which converges
asymptotically to the true posterior.  We discuss additional differences between
our model and the DILN-HMM in Sec. \ref{sec:prom-local-trans}.

One class of application in which it is useful to incorporate a notion of locality 
occurs when the latent state sequence consists of several parallel
chains, so that the global state changes incrementally, but where
these increments are not independent across chains.  
Factorial HMMs \cite{ghahramani1997factorial} are commonly used in
this setting, but this ignores dependence among chains, and hence may do poorly when some
combinations of states are much more probable than suggested by the
chain-wise dynamics.

Another setting where the LT property is useful is when there is a
notion of state geometry that licenses syllogisms: e.g., if A
frequently leads to B and C and B frequently leads to D and E, 
then it may be sensible to infer that A and C may lead
to D and E as well.  This property is arguably present in musical harmony, where 
consecutive chords are often (near-)neighbors in the ``circle of
fifths'', and small steps along the circle are more
common than large ones.

The paper is structured as follows: In section \ref{sec:model} we
define the model.  In section \ref{sec:inference}, we develop a
Gibbs sampling algorithm 
based on an augmented data representation, which we call the Markov
Jump Process with Failed Transitions (MJP-FT).  In section
\ref{sec:experiments} we test two versions of the model: one
on a speaker diarization task in which the speakers are
inter-dependent, and another on a four-part chorale corpus,
demonstrating performance improvements over state-of-the-art 
models when ``local transitions''
are more common in the data.  Using sythetic data from an HDP-HMM, we
show that the LT variant can learn not to use its similarity bias when
the data does not support it.  Finally, in section \ref{sec:discussion}, we 
conclude and discuss the relationships between the HDP-HMM-LT and
existing HMM variants.  Code and additional details are available
at \url{http://colindawson.net/hdp-hmm-lt/}

\section{An HDP-HMM With Local Transitions}
\label{sec:model}
We wish to add to the transition model the concept of a transition to
a ``nearby'' state, where transitions between states $j$ and $j'$ are
more likely {\em a priori} to the extent that they are ``nearby'' in
some similarity space.  In order to accomplish this, we first
consider an alternative construction of the transition distributions,
based on the Normalized Gamma Process representation of the DP
\cite{ishwaran2002exact, ferguson1973bayesian}.

\subsection{A Normalized Gamma Process representation of the HDP-HMM}
\label{sec:normalized-gamma}

The Dirichlet Process is an instance of a normalized completely random measure
\cite{kingman1967completely, ferguson1973bayesian},
that can be defined as $G = \sum_{k=1}^{\infty} \tilde{\pi}_k \delta_{\theta_k}$, where 
\begin{align}
  \pi_{k} \stackrel{\text{ind.}}{\sim} \Gamm{\alpha \beta_{k}}{1} \quad T =
  \sum_{k=1}^{\infty} \pi_{k} \quad
  \tilde{\pi}_{k} = \frac{\pi_{k}}{T} \label{eq:20},
\end{align}
$\delta_{\theta}$ is a measure assigning 1 to sets if they contain $\theta$ and
0 otherwise, and subject to the constraint 
that $\sum_{k\geq 1} \beta_{k} = 1$ and $0 < \alpha < \infty$.  It
has been shown \cite{ferguson1973bayesian, paisley2012discrete, favaro2013mcmc}
that the normalization constant $T$ is positive and finite almost
surely, and that $G$ is distributed as a DP
with base measure $G_0 = \sum_{k=1}^{\infty} \beta_{k} \delta_{\theta_{k}}$.
If we draw $\bbeta = (\beta_1, \beta_2, \dots)$ from the
$\GEM{\gamma}$ stick-breaking process, draw an
i.i.d.~sequence of $\theta_k$ from a base measure $H$, and then draw an 
i.i.d.~sequence of random measures, $\{G_j\}, j = 1, 2, \dots$, from the above process,
this defines a Hierarchical Dirichlet Process (HDP).  If each $G_j$
is associated with the hidden states of an HMM, $\bpi$ is the infinite
matrix where entry $\pi_{jj'}$ is
the $j'$th mass associated with the $j$th random measure, and
$T_j$ is the sum of row $j$, then we obtain the prior for the HDP-HMM, where
\begin{align}
  \label{eq:50}
  p(z_t \given z_{t-1}, \bpi) = \tilde{\pi}_{z_{t-1}z_t} = \pi_{jj'} / T_j
\end{align}

\subsection{Promoting ``Local'' Transitions}
\label{sec:prom-local-trans}

In the HDP prior, the rows of the transition matrix 
are conditionally independent.  We wish to relax this
assumption, to incorporate possible prior knowledge
that certain pairs of states are ``nearby'' in some sense and thus 
more likely than others to produce large transition weights between
them (in both directions); that is, transitions are likely to be ``local''.
We accomplish this by associating each latent state $j$ with a location  
$\ell_j$ in some space $\Omega$, introducing a ``similarity function'' $\phi: \Omega   
\times \Omega \to (0,1]$, and scaling each element
$\pi_{jj'}$ by $\phi_{jj'} = \phi(\ell_{j}, \ell_{j'})$.  For  
example, we might wish to define a (possibly asymmetric) divergence function $d:
\Omega \times \Omega \to [0,\infty)$ and set $\phi(\ell_j,\ell_j)
= \exp\{-d(\ell_j,\ell_{j'})\}$ so that transitions are less likely the
farther apart two states are.  By setting $\phi \equiv 1$, we
obtain the standard HDP-HMM. 
The DILN-HMM \cite{zhu2016hidden},
employs a similar
rescaling of transition probabilities via an exponentiated Gaussian
Process, following \cite{paisley2012discrete},
but the scaling function must be positive semi-definite, and in particular
symmetric, whereas in the HDP-HMM-LT, $\phi$
need only take values in $(0,1]$.  Moreover, the DILN-HMM does not
allow the scales to be tied to other state parameters, and hence
encode an independent notion of similarity.  

Letting $\bell = (\ell_1, \ell_2, \dots)$, we can replace
\eqref{eq:20} for $j \geq 1$ by
\begin{align}
\begin{split}
  &\pi_{jj'} \given \bbeta, \bell \sim \Gamm{\alpha
    \beta_{j'}}{1}, \quad
  T_j = \sum_{j'=1}^{\infty} \pi_{jj'}\phi_{jj'} \\
&\tilde{\pi}_{jj'} = \pi_{jj'}\phi_{jj'} / T_j, \quad p(z_t \given
z_{t-1}, \bpi, \bell) = \tilde{\pi}_{z_{t-1}z_t}.
\end{split}
\end{align}
Since the $\phi_{jj'}$ are positive and
bounded above by 1, 
\begin{equation}
0 < \pi_{j1}\phi_{j1} \leq T_j \leq \sum_{j'} \pi_{jj'} < \infty
\end{equation}
almost surely, where the last inequality carries over from the original
HDP. The prior means of the unnormalized transition distributions, $\bpi_j$ are then 
proportional (for each $j$) to $\alpha\bbeta\bphi_{j}$ where $\bphi_j
= (\phi_{j1}, \phi_{j2}, \dots)$.

The distribution of the latent state sequence
$\bz$ given $\bpi$ and $\bell$ is now
\begin{align}
\begin{split}
  p(\bz \given \bpi, \bell) &= \prod_{t=1}^T \pi_{z_{t-1}z_t}
  \phi_{z_{t-1}z_t} T_{z_{t-1}}^{-n_{z_{t-1}\cdot}} \\
  &= \prod_{j=1}^\infty T_j^{-1} \prod_{j'=1}^{\infty}
  \pi_{jj'}^{n_{jj'}} \phi_{jj'}^{n_{jj'}}
\end{split}
\end{align}
where $n_{jj'} = \sum_{t=1}^T I(z_{t-1} = j, z_{t} = j')$ is the
number of transitions from state $j$ to state $j'$ in the sequence
$\bz$ and $n_{j\cdot} = \sum_{j'} n_{jj'}$ is the total number of
visits to state $j$.  Since $T_j$ is a sum over products of
$\pi_{jj'}$ and $\phi_{jj'}$ terms, the posterior for
$\bpi$ is no longer a DP.  However, conditional conjugacy can be
restored by a data-augmentation process with a natural interpretation, 
which is described next.

\subsection{The HDP-HMM-LT as the Marginalization of
a Markov Jump Process with ``Failed'' Transitions}
\label{sec:mjp-ft}

In this section, we define a stochastic process that we call the
Markov Jump Process with Failed Transitions (MJP-FT), from which we obtain the HDP-HMM-LT
by marginalizing over some of the variables.  By reinstating these
auxiliary variables, we obtain a simple Gibbs sampling algorithm over
the full MJP-FT, which can be used to sample from the marginal
posterior of the variables used by the HDP-HMM-LT.

Let $\bbeta$, $\bpi$, $\bell$ and $T_j, j = 1, 2, \dots$ be defined as
in the last section. Consider a continuous-time Markov Process over the states $j = 1, 2,
\dots$, and suppose that if the process makes a jump to state 
$z_{t}$ at time $\tau_t$, the next jump, which is to state $z_{t+1}$, occurs at
time $\tau_t + \tilde{u}_t$, where $\tilde{u}_{t} \sim \Exp{\sum_{j'} \pi_{jj'}}$,
and $p(z_{t+1} = j' \given z_{t} = j) \propto
\pi_{jj'}$, independent of $\tilde{u}_t$.  Note that in this
formulation, unlike in standard formulations of Markov Jump Processes,
we are assuming that self-jumps are possible.

If we only observe the jump sequence $\bz$ and not the holding times
$\tilde{u}_t$, this is an ordinary Markov chain with transition matrix row-proportional to
$\bpi$.  If we do not observe the jumps directly, but
instead an observation is generated once per jump from a distribution that depends
on the state being jumped to, then we have an ordinary HMM whose
transition matrix is obtained by normalizing $\bpi$; that
is, we have the HDP-HMM.

We modify this process as follows.  
Suppose each jump attempt from state $j$ to state $j'$ has probability
$(1 - \phi_{jj'})$ of failing, in which case no transition occurs and
no observation is generated.  Assuming independent failures, the rates of 
successful and failed jumps from $j$ to $j'$ are $\pi_{jj'}\phi_{jj'}$ and
$\pi_{jj'}(1-\phi_{jj'})$, respectively. 
The probability that the first successful jump is to
state $j'$ (that is, that $z_{t+1} = j'$) 
is proportional to the rate of successful jump attempts to $j'$, which
is $\pi_{jj'}\phi_{jj'}$.  Conditioned on $z_t$, the holding time, $\tilde{u}_{t}$, is
independent of $z_{t+1}$ and is distributed as
$\mathsf{Exp}(T_{z_t})$.  We denote the
total time spent in state $j$ by $u_j = \sum_{t: z_t = j}
\tilde{u}_t$, where, as the sum of i.i.d.~Exponentials,
\begin{equation}
u_j \given \bz, \bpi, \btheta \stackrel{\text{ind.}}{\sim} \Gamm{n_{j\cdot}}{T_j}
\end{equation}
During this period there will be $q_{jj'}$
failed attempts to jump to state $j'$, where $q_{jj'} \sim \Pois{u_j
\pi_{jj'}(1 - \phi_{jj'})}$ are independent.  This data augmentation
bears some conceptual similarity to the Geometrically distributed
$\rho$ auxiliary variables introduced to the
HDP-HSMM \cite{johnson2013bayesian} to restore conditional conjugacy.
However, there are key differences: first, $\rho$ measure how many steps the chain
would have remained in state j under Markovian dynamics, whereas our
$u$ represents putative continuous holding times between each
transition, and second $\rho$ allows for the restoration of a 
zeroed out entry in each row, whereas $u$ allows us to work with
unnormalized $\pi$ entries, avoiding the need to restore zeroed out
entries in the HSMM-LT

Incorporating
$\bu = \{u_j\}$ and $\bQ = \{q_{jj'}\}$ as augmented data simplifies 
the likelihood for $\bpi$, yielding
\begin{align}
  p(\bz, &\bu, \bQ \given \bpi) =  p(\bz \given \bpi) p(\bu \given
  \bz, \bpi) p(\bQ \given \bu, \bpi)
\end{align}
where dependence on $\bell$ has been omitted for conciseness.  After
grouping terms and omitting terms that do not depend on $\bpi$, this
proportional (as a function of $\bpi$) to
\begin{align}
\begin{split}
\label{eq:joint-likelihood}
\prod_{j} \prod_{j'} \pi_{jj'}^{n_{jj'} + q_{jj'}} \phi_{jj'}^{n_{jj'}}
  (1-\phi_{jj'})^{q_{jj'}}
  e^{-\pi_{jj'}u_j}
\end{split}
\end{align}
Conveniently, the $T_j$ have canceled, and the exponential terms involving
$\pi_{jj'}$ and $\phi_{jj'}$ in the Gamma and Poisson distributions of
$u_{j}$ and $q_{jj'}$ combine to cause $\phi_{jj'}$ to vanish.

Additional details and derivations for this data augmentation are in Appendix \ref{sec:mjp-ft}.

\subsection{Sticky and Semi-Markov Generalizations}
\label{sec:an-hsmm-modification}

We note that the local transition property of
the HDP-HMM-LT can be combined with the Sticky property of the Sticky
HDP-HMM \cite{fox2008hdp}, or the non-geometric duration distributions of the
HDP-HSMM \cite{johnson2013bayesian}, to add
additional prior weight on self-transitions.  In the
former case, no changes to inference are needed; one can simply add the
the extra mass $\kappa$ to the shape parameter of the Gamma prior on
the $\pi_{jj}$, and employ the same auxiliary variable method used by
\citeauthor{fox2008hdp} to distinguish ``Sticky'' from ``regular'' self-transitions.
For the semi-Markov case, we can fix the diagonal
elements of $\bpi$ to zero, and allow $D_t$ observations to be
emitted $i.i.d.$ according to a state-specific duration distribution, and sample the
latent state sequence using a suitable semi-Markov message passing algorithm
\cite{johnson2013bayesian}.  Inference for the $\bphi$ matrix
is not affected, since the diagonal elements are assumed to be 1.
Unlike in the original representation of the HDP-HSMM, no further
data-augmentation is needed, as the (continuous) durations $\bu$ already
account for the normalization of the $\bpi$.

\subsection{Obtaining the Factorial HMM as a Limiting Case}

One setting in which a local transition property is
desirable is the case where the latent states encode
multiple hidden features at time $t$ as a vector of categories. 
Such problems are often modeled using factorial HMMs \cite{ghahramani1997factorial}.  In
fact, the HDP-HMM-LT yields the factorial HMM in the limit as
$\alpha,\gamma \to \infty$, fixing each row of $\pi$ to be uniform 
with probability 1, so the dynamics are controlled entirely by
$\phi$.  If $\mathbf{A}^{(d)}$ is the transition matrix for chain $d$, then
setting $\phi(\bell_j, \bell_{j'}) =
\exp{-d(\bell_j,\bell_{j'})}$ with asymmetric ``divergences'' 
$d(\bell_j,\bell_{j'}) = -\sum_d \log(\mathbf{A}^{(d)}_{\ell_{jd},\ell_{j'd}})$
yields the factorial transition model.

\subsection{An Infinite Factorial HDP-HMM-LT}

Nonparametric extensions of the factorial HMM, such as the infinite factorial
hidden Markov Model \cite{gael2009infinite} and the infinite factorial dynamic model
\cite{valera2015infinite}, have been developed in recent years
by making use of the Indian Buffet Process
\cite{ghahramani2005infinite} as a state prior.  It would be
conceptually straightforward to combine the IBP state prior with the
similarity bias of the LT model, provided the chosen similarity
function is uniformly bounded above on the space of infinite length
binary vectors (for example, take $\phi(u,v)$ to be the exponentiated
negative Hamming distance between $u$ and $v$).  Since the number of
differences between two draws from the IBP is finite with probability
1, this yields a reasonable similarity metric.

\section{Inference}
\label{sec:inference}

We develop a Gibbs sampling algorithm based on the MJP-FT
representation described in Sec. \ref{sec:mjp-ft}, augmenting the data with the duration
variables $\bu$, the failed jump attempt count matrix, $\bQ$, as well
as additional auxiliary variables which we will define below.
In this representation the transition matrix is not represented
directly, but is a deterministic function of the unscaled transition
``rate'' matrix, $\bpi$, and the similarity matrix, $\bphi$.  
The full set of variables is partitioned into blocks: $\{\gamma, \alpha, \beta, \bpi\}$,
$\{\bz, \bu, \bQ, \Lambda\}$, $\{\theta, \bell\}$, and $\{\xi\}$, where $\Lambda$
represents a set of auxiliary variables that will be introduced
below, $\theta$ represents the emission
parameters (which may be further blocked depending on the specific
choice of model), and $\xi$ represents additional
parameters such as any free parameters of the similarity function,
$\phi$, and any hyperparameters of the emission distribution.

\subsection{Sampling Transition Parameters and Hyperparameters}

The joint posterior over $\gamma$, $\alpha$, $\bbeta$ and $\bpi$
given the augmented data $\mathcal{D} = (\bz, \bu, \bQ, \Lambda)$ 
will factor as
\begin{align}
  \label{eq:46}
  \begin{split}
  p(\gamma, &\alpha, \bbeta, \bpi \given \mathcal{D}) \\&= p(\gamma
  \given \mathcal{D}) p(\alpha \given \mathcal{D}) p(\beta \given \gamma, \mathcal{D}) p(\bpi
  \given \alpha, \beta, \mathcal{D})
\end{split}
\end{align}
We describe these four factors in reverse order.  For additional
details, see Appendix \ref{sec:hdp-inference}.

\paragraph{Sampling \texorpdfstring{$\bpi$}{pi}} 
Having used data augmentation to simplify the likelihood for $\bpi$ to
the factored conjugate form in \eqref{eq:joint-likelihood}, the individual
$\pi_{jj'}$ are {\it a posteriori} independent $\Gamm{\alpha\beta_{j'} + n_{jj'} + q_{jj'}}{1 + u_j}$ distributed.

\paragraph{Sampling \texorpdfstring{$\bbeta$}{beta}}
\label{sec:sampling-bbeta}
To enable joint sampling of $\bz$, we employ a
weak limit approximation to the HDP \cite{johnson2013bayesian}, approximating the stick-breaking
process for $\bbeta$ using a finite Dirichlet distribution with a
$J$ components, where $J$ is larger than we expect to
need.  Due to the product-of-Gammas form, we can integrate out $\bpi$ 
analytically to obtain the marginal likelihood:
\begin{align}
\label{eq:beta-prior-likelihood}
  &p(\beta \given \gamma) = \frac{\Gamma(\gamma /
    J)^J}{\Gamma(\gamma)} \prod_{j} \beta_{j}^{\frac{\gamma}{J} - 1}
  \\
  &p(\mathcal{D} \given \beta, \alpha) \propto
  \prod_{j=1}^J (1+u_j)^{-\alpha} \prod_{j'}
    \frac{\Gamma(\alpha\beta_{j'} + n_{jj'} +
    q_{jj'})}{\Gamma(\alpha\beta_{j'})} \notag
\end{align}
 where we have used the fact that the $\beta_j$ sum to 1 to pull out
 terms of the form $(1 + u_j)^{-\alpha\beta_{j'}}$ from the inner
 product in the likelihood.  Following \citet{teh2006hierarchical}, 
we can introduce auxiliary variables $\bM = \{m_{jj'}\}$, with
\begin{equation}
  \label{eq:m-distribution}
  p(m_{jj'} \given \beta_{j'}, \alpha, \mathcal{D}) \stackrel{ind}{\propto}
  s_{n_{jj'} + q_{jj'}, m_{jj'}} \alpha^{m_{jj'}}
    \beta_{j'}^{m_{jj'}}
\end{equation}
for integer $m_{jj'}$ ranging between $0$ and $n_{jj'} + q_{jj'}$,
where $s_{n,m}$ is an unsigned Stirling number of the first kind.
The normalizing constant in this distribution cancels the ratio of
Gamma functions in the $\bbeta$ likelihood, so, letting $m_{\cdot j'} =
\sum_{j} m_{jj'}$ and $m_{\cdot\cdot} = \sum_{j'} m_{\cdot j'}$, 
the posterior for (the truncated) $\bbeta$ is a Dirichlet
whose $j$th mass parameter is $\frac{\gamma}{J} + m_{\cdot j}$.

\paragraph{Sampling Concentration Parameters}
\label{sec:sampling-alpha}
Incorporating $\bM$ into $\mathcal{D}$, we can integrate out
$\bbeta$ to obtain
\begin{align}
\begin{split}
\label{eq:gamma-marginal-likelihood}
  p(\mathcal{D} \given \alpha, \gamma) &\propto
  \alpha^{m_{\cdot\cdot}} e^{-\sum_{j''} \log(1+u_{j''}) \alpha}
  \\
  & \qquad \frac{\Gamma(\gamma)}{\Gamma(\gamma + m_{\cdot\cdot})}\times \prod_j
  \frac{\Gamma(\frac{\gamma}{J} + m_{\cdot
      j})}{\Gamma(\frac{\gamma}{J}) }
\end{split}
\end{align}
Assuming that $\alpha$ and $\gamma$ have Gamma priors with shape and
rate parameters $a_{\alpha}, b_{\alpha}$ and $a_{\gamma}, b_{\gamma}$,
then
\begin{equation}
  \label{eq:alpha-posterior}
  \alpha \given \mathcal{D} \sim \Gamm{a_{\alpha}
    + m_{\cdot\cdot}}{b_\alpha + \sum_j\log(1+u_j)}.
\end{equation}
To simplify the likelihood for $\gamma$, we can introduce a final
set of auxiliary variables, $\br = (r_1, \dots,
r_J)$, $r_{j'} \in \{0,\dots,m_{\cdot j'}\}$ and $w \in (0,1)$ with the following distributions:
\begin{align}
  \label{eq:9}
  &p(r_{j'}  = r \given m_{\cdot {j'}}, \gamma) \propto s(m_{\cdot {j'}}, r)
    \left(\frac{\gamma}{J}\right)^r \\
  &p(w \given m_{\cdot\cdot} \gamma) \propto 
    w^{\gamma - 1} (1-w)^{m_{\cdot\cdot} - 1}
\end{align}
The normalizing constants are ratios of Gamma functions, which cancel
those in \eqref{eq:gamma-marginal-likelihood}, so that
\begin{equation}
  \label{eq:18}
  \gamma \given \mathcal{D},\br, w \sim \Gamm{a_{\gamma} + r_{\cdot}}{b_{\gamma} - \log(w)}
\end{equation}

\subsection{Sampling \texorpdfstring{$\bz$}{the latent state sequece} and the auxiliary variables}
\label{sec:sampling-z_t}

We sample the hidden state sequence, $\bz$, jointly with the auxiliary
variables, which consist of $\bu$, $\bQ$, $\bM$, $\br$ and $w$.  The
joint conditional distribution of these variables is defined directly
by the generative model:
\begin{align*}
  p(\mathcal{D}) = p(\bz) p(\bu \given \bz) p(\bQ \given
  \bu) p(\bM \given \bz, \bQ) p(\br \given \bM) p(w
  \given\bM)
\end{align*}
Since we are conditioning on the transition matrix, we can
sample the entire sequence $\bz$ jointly with the forward-backward algorithm,
as in an ordinary HMM. Since we are sampling
the labels jointly, this step requires $\mathcal{O}(TJ^2)$
computation per iteration, which is the bottleneck of the inference algorithm for
reasonably large $T$ or $J$ (other updates are constant in $T$ or in
$J$). Having done this, we can sample $\bu$, $\bQ$, $\bM$,
$\br$ and $w$ from their forward distributions.  It is also possible to employ a variant on beam sampling 
\cite{vangael2008beam} to speed up each iteration,
at the cost of slower mixing, but we did not use this variant here.

\subsection{Sampling state and emission parameters}
\label{sec:sampling-eta}

Depending on the application, the locations $\bell$ may or may not
depend on the emission parameters, $\btheta$.  If not, 
sampling $\btheta$ conditional on $\bz$ is
unchanged from the HDP-HMM.  There is no general-purpose method
for sampling $\bell$, or for sampling $\btheta$ in the dependent case,
due to the dependence on the form of $\phi$ and on the emission model,
but specific instances are illustrated in the experiments below.

\section{Experiments}
\label{sec:experiments}

The parameter space for the hidden states, the associated prior $H$ on
$\btheta$, and the similarity function
$\phi$, is application-specific; we consider here two cases.  The
first is a speaker-diarization task, where 
each state consists of a finite $D$-dimensional binary
vector whose entries indicate which speakers are currently
speaking.  In this experiment, the state vectors both determine
the pairwise similarities and partially determine the emission 
distributions via a linear-Gaussian model.
In the second experiment, the data consists of Bach chorales, and the
latent states can be thought of as harmonic contexts.  There, the components
of the states that govern similarities are modeled as 
independent of the emission distributions, which are categorical 
distributions over four-voice chords.

\subsection{Cocktail Party}

\begin{figure}[tb]
\centering
  \centerline{\includegraphics[width = \columnwidth, height = 0.11\columnwidth]{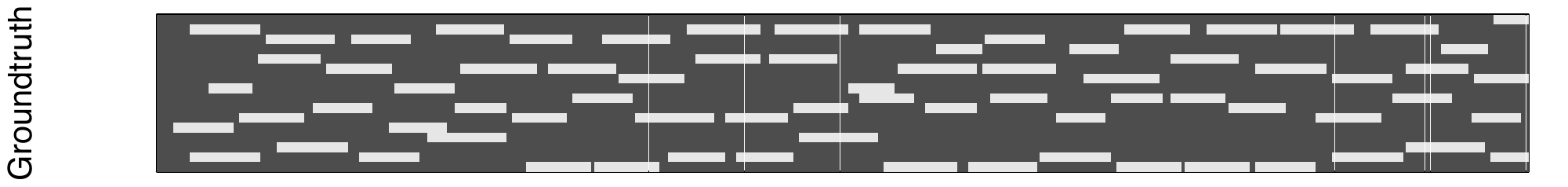}}
  \centerline{\includegraphics[width = \columnwidth, height = 0.11\columnwidth]{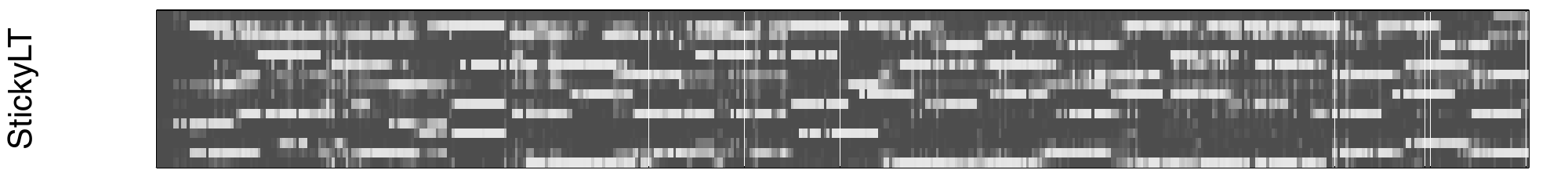}}
  \centerline{\includegraphics[width = \columnwidth, height = 0.11\columnwidth]{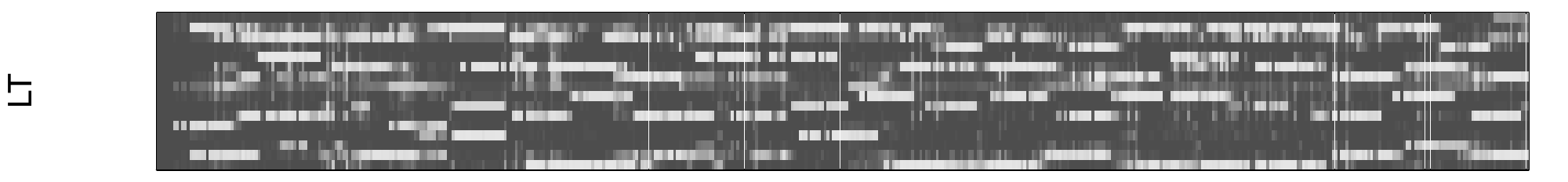}}
  \centerline{\includegraphics[width = \columnwidth, height = 0.11\columnwidth]{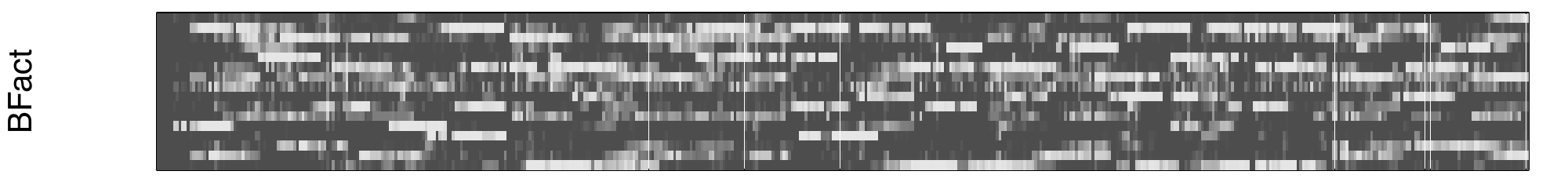}}
  \centerline{\includegraphics[width = \columnwidth, height = 0.11\columnwidth]{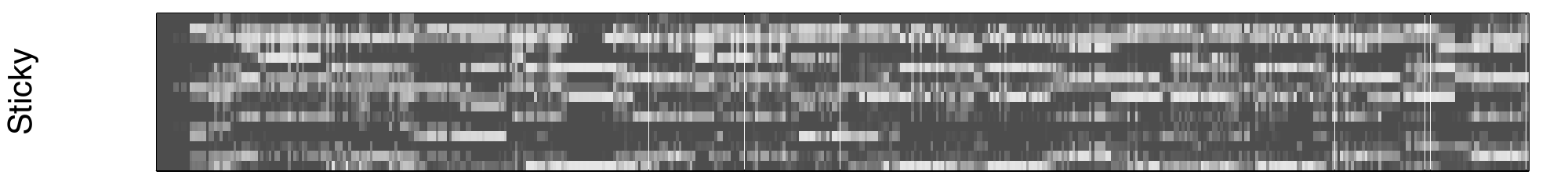}}
  \centerline{\includegraphics[width = \columnwidth, height =
    0.11\columnwidth]{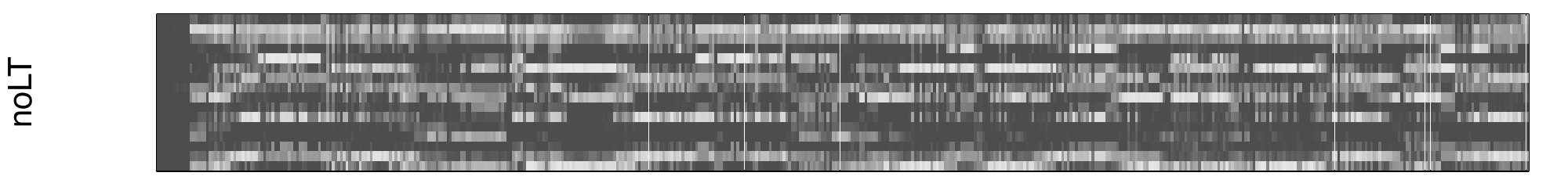}}
\caption{Binary speaker matrices for the cocktail data, with time on
  the horizontal axis and speaker on the vertical axis.  White is 1,
  black is 0.  The ground truth matrix is at the top, followed by the 
  inferred speaker matrix for the
  Sticky HDP-HMM-LT, HDP-HMM-LT, binary factorial, Sticky-HDP-HMM, and
  ``vanilla'' HDP-HMM.  All inferred matrices are averaged over 5 runs
  of 5000 Gibbs iterations each, with the first 2000 iterations
  discarded as burn-in. \label{fig:cocktail-binary-matrices}}
\end{figure}

\paragraph{The Data} The data was constructed using audio signals
collected from the PASCAL 1st Speech Separation
Challenge\footnote{\scriptsize \url{http://laslab.org/SpeechSeparationChallenge/}}.
The underlying signal consisted of $D = 16$ speaker channels recorded
at each of $T = 2000$ time steps, with the resulting $T \times D$
signal matrix, denoted by $\btheta^{*}$, mapped to $K = 12$ microphone
channels via a weight matrix, $\bW$.  
The 16 speakers were grouped into 4 conversational groups of 4, 
where speakers within a conversation took turns
speaking (see Fig. \ref{fig:cocktail-binary-matrices}).   
In such a task, there are naively $2^{D}$ possible
states (here, $65536$).  However, due to the conversational grouping, if
at most one speaker in a conversation is speaking at any given
time, the state space is constrained, with only $\prod_c (s_c + 1)$
states possible, where $s_c$ is the number of speakers in conversation
$c$ (in this case $s_c \equiv 4$, for a total of 625
possible states).

Each ``turn'' within a conversation consisted of a single sentence
(average duration $\sim 3$s) and turn orders within a conversation were
randomly generated, with random pauses distributed as $\Norm{1/4s}{(1/4s)^2}$
inserted between sentences.  Every time a speaker has a turn, the
sentence is drawn randomly from the 500 sentences uttered by that
speaker in the data.  The conversations continued for 40s, and 
the signal was down-sampled to length 2000.  The 'on' portions of each
speaker's signal were normalized to have amplitudes with mean 1 and standard
deviation 0.5.  An additional column of 1s was added to the speaker
signal matrix, $\btheta^{*}$,
representing background noise.  
The resulting signal matrix, denoted $\btheta^{*}$, was thus
$2000 \times 17$ and the weight matrix, $\bW$, was $17 \times 12$.  Following
\citet{gael2009infinite} and \citet{valera2015infinite}, the weights
were drawn independently from a $\Unif{0,1}$ distribution, and
independent $\Norm{0}{0.3^2}$ noise was added to each entry of the
observation matrix.

\paragraph{The Model}
The latent states, $\btheta_j$, are the $D$-dimensional
binary vectors whose $d$th entry indicates whether or not speaker $d$ is
speaking.  The locations $\bell_j$ are identified with the binary
vectors, $\bell_j := \btheta_j$.  We use a Laplacian similarity function
on Hamming distance, $d_0$, so that
$\phi_{jj'} := \exp(-\lambda d_{0}(\bell_j,
\bell_{j'})), \lambda \geq 0$.  
The emission model is linear-Gaussian as in the data, with $(D + 1)
\times K$ weight matrix $\bW$, and $T \times (D + 1)$ signal matrix
$\btheta^*$ whose $t^{\rm th}$ row is $\btheta_t := (1, \btheta_{z_t})$, so that 
$\by_t \given \bz \sim \Norm{\bW^{\sf T} \btheta^{*}_{t}}{\bSigma}$.  
For the experiments discussed here, we assume that $\bSigma$ 
is independent of $j$, but this assumption
is easily relaxed if appropriate.  

For finite-length binary vector
states, the set of possible states is finite, and so it may
seem that a nonparametric model is unnecessary.  However, if $D$ is
reasonably large, likely most of the $2^D$ possible states
are vanishingly unlikely (and the number of observations may
well be less than $2^D$), and so we would like to encourage
the selection of a sparse set of states.  Moreover, there could be more
than one state with the same emission parameters, but with different transition
dynamics.  Next we describe the additional inference steps needed for
this version of the model.

\paragraph{Sampling $\btheta$ / $\bell$} Since $\btheta_j$ and
$\bell_j$ are identified, influencing both the
transition matrix and the emission distributions, both the state
sequence $\bz$ and the observation matrix $\bY$
are used in the update.  We put independent Beta-Bernoulli priors
on each coordinate of $\btheta$, and Gibbs sample each coordinate $\theta_{jd}$
conditioned on all the others and the coordinate-wise prior means,
$\{\mu_d\}$, which we sample in turn conditioned on $\btheta$.
Details are in Appendix \ref{sec:cocktail-inference}.

\paragraph{Sampling \texorpdfstring{$\lambda$}{kernel decay rate}}
\label{sec:sampling-lambda}
The $\lambda$ parameter of the similarity function governs the
connection between $\bell$ and $\bphi$.  Substituting the definition
of $\bphi$ into \eqref{eq:joint-likelihood} yields
\begin{equation}
  \label{eq:88}
  p(\bz, \bQ \given \bell, \lambda) \propto \prod_{j}\prod_{j'}
  e^{-\lambda d_{jj'} n_{jj'}}(1-e^{-\lambda d_{jj'}})^{q_{jj'}} 
\end{equation}
We put an $\Exp{b_{\lambda}}$ prior on $\lambda$, which yields a
posterior density
\begin{align}
  \label{eq:88}
  p(\lambda \given \bz, \bQ, \bell) &\propto
  e^{-(b_{\lambda} + \sum_{j}\sum_{j'} d_{jj'} n_{jj'})\lambda}
  \\ & \qquad \times \prod_{j}\prod_{j'}
  (1-e^{-\lambda d_{jj'}})^{q_{jj'}} \notag
\end{align}
This density is log-concave, and so we use Adaptive Rejection Sampling \cite{gilks1992adaptive}
to sample from it.

\paragraph{Sampling \texorpdfstring{$\bW$}{weights} and
  \texorpdfstring{$\bSigma$}{emission covariance}}
Conditioned on $\bY$ and $\btheta^*$, 
$\bW$ and $\bSigma$ can be sampled as in
Bayesian linear regression.  If each column of $\bW$ has a
multivariate Normal prior, then the columns are {\em a posteriori} independent
multivariate Normals.  For the experiments
reported here, we fix $\bW$ to its ground truth value so that 
$\btheta^*$ can be compared directly with the ground truth signal matrix, and we 
constrain $\bSigma$ to be diagonal, with Inverse Gamma priors on the 
variances, resulting in conjugate updates.

\begin{figure}[tb]
\centering
  \centerline{\includegraphics[width = 0.72\columnwidth, height = 1.8cm]{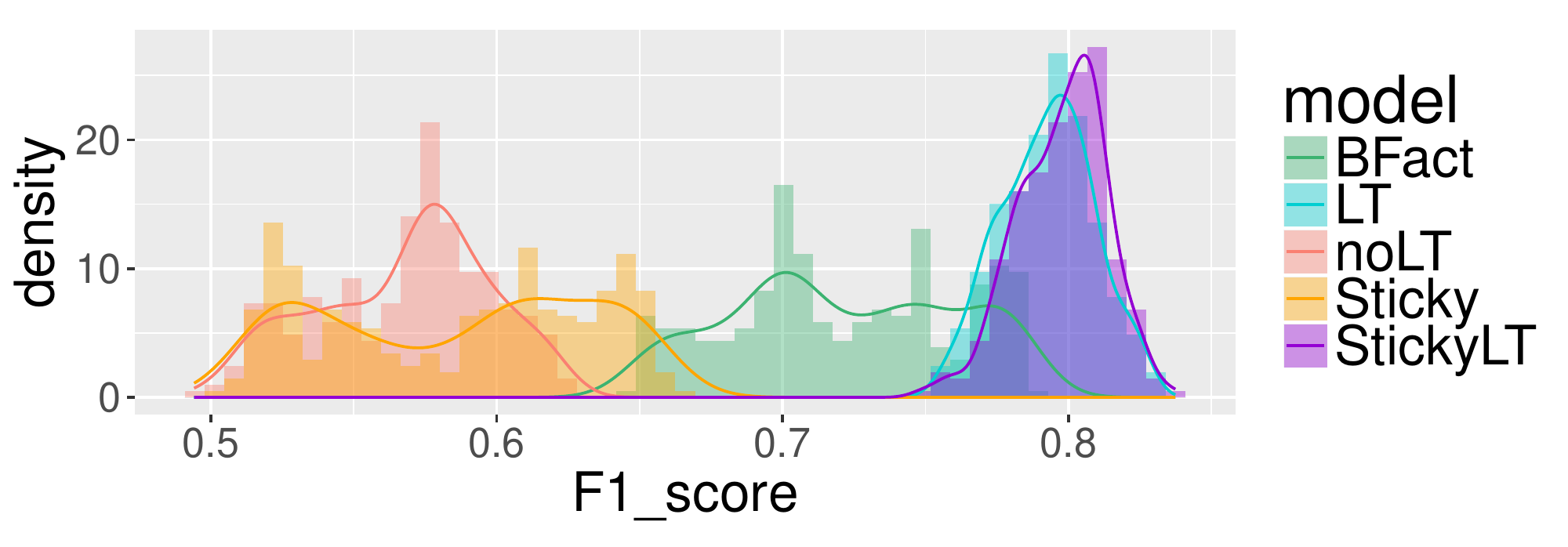}}
  \centerline{\includegraphics[width = 0.72\columnwidth, height = 1.8cm]{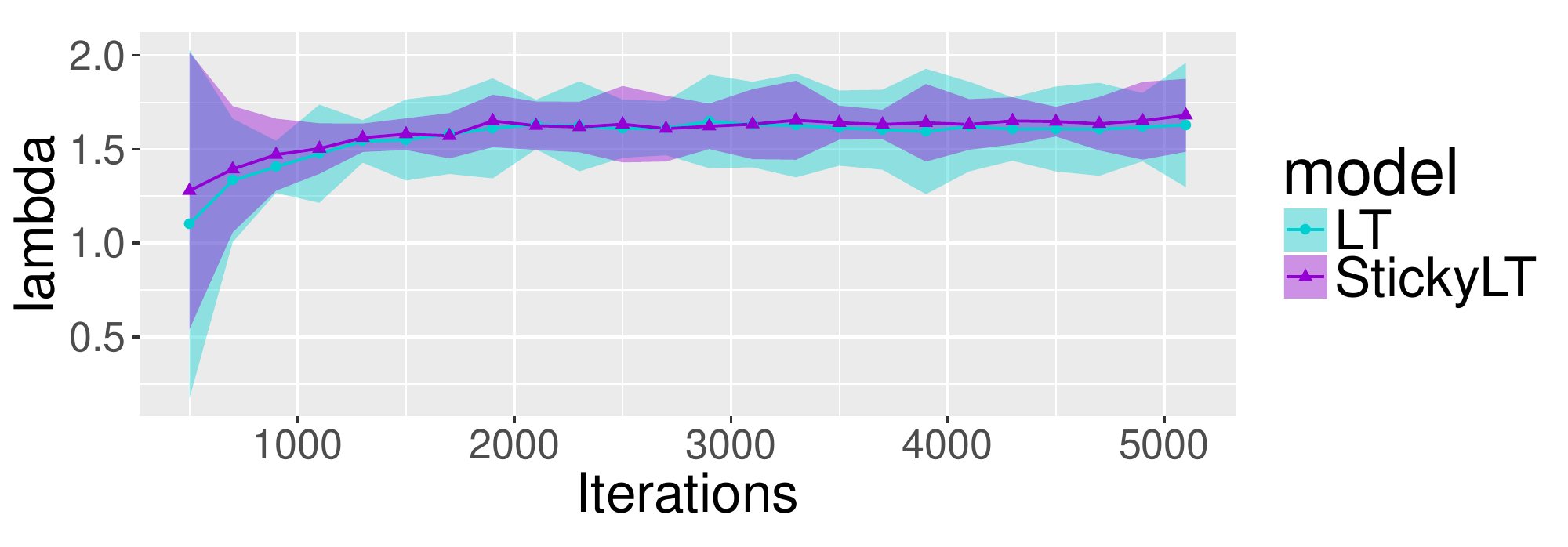}}
  \centerline{\includegraphics[width = 0.72\columnwidth, height =
    1.8cm]{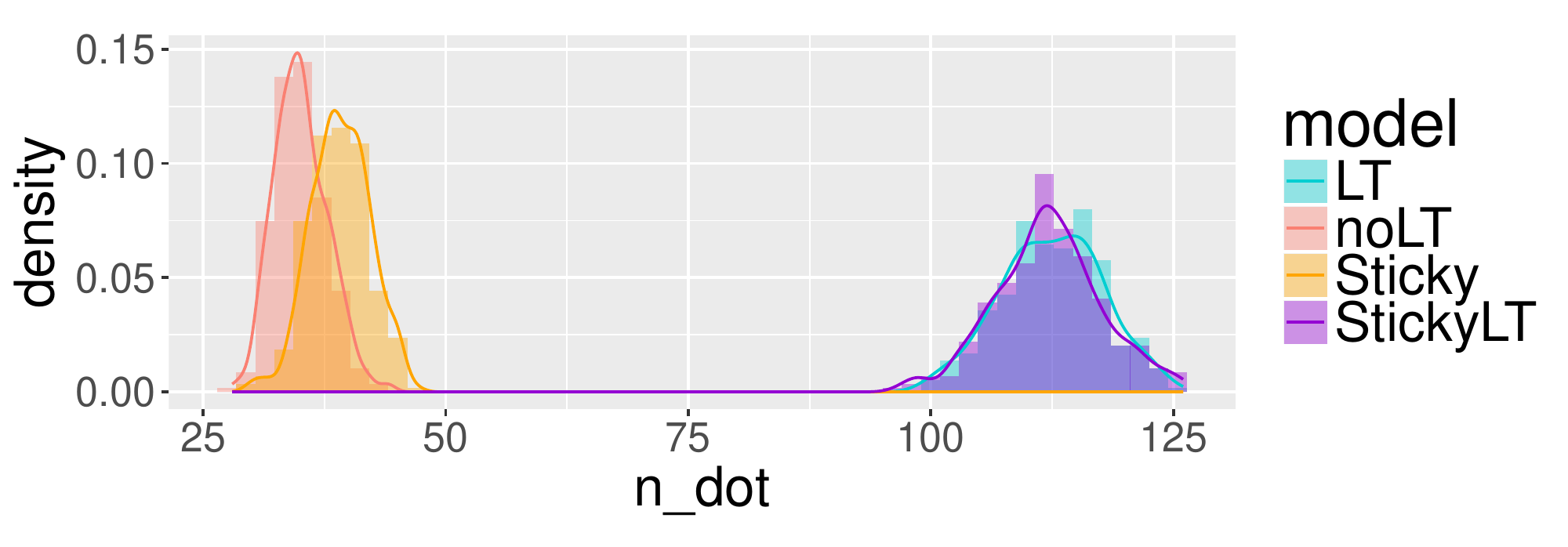}}
\vspace{-0.2cm}
\caption{Top: F1 score for inferred relative to ground
  truth binary speaker matrices on cocktail party data, evaluated 
  every 50th Gibbs iteration after the first 2000, aggregating across 5
  runs of each model. 
  Middle: Inferred $\lambda$, for the LT and Sticky-LT
  models by Gibbs iteration, averaged over 5 runs.  Bottom: Number of
  states used, $n_{\cdot}$, by each model in the training set. Error bands are
  99\% confidence interval of the mean per iteration. \label{fig:cocktail-metrics}
}
\end{figure}

\paragraph{Results}
We attempted to infer the binary speaker matrices
using five models: (1) a binary-state Factorial HMM \cite{ghahramani1997factorial}, where
individual binary speaker sequences are modeled as independent, 
(2) an ordinary HDP-HMM without local transitions \cite{teh2006hierarchical}, where the
latent states are binary vectors, (3) a Sticky HDP-HMM \cite{fox2008hdp}, (4) our
HDP-HMM-LT model, and (5) a model that combines the Sticky and LT
properties\footnote{We attempted to add a comparison to the DILN-HMM
  \cite{zhu2016hidden} as well, but code could not be obtained, 
  and the paper did not provide enough detail to reproduce their inference algorithm.}.
For all models, all concentration and noise 
precision parameters are given $\Gamm{0.1}{0.1}$ priors.  For the
Sticky models, the ratio $\frac{\kappa}{\alpha + \kappa}$ is given a
$\Unif{0,1}$ prior. We evaluated the models at each
iteration using both the Hamming distance between inferred 
and ground truth state matrices and F1 score.  We also plot the inferred decay
rate $\lambda$, and the number of states used by the
LT and Sticky-LT models.  The results for the five models are in Fig.
\ref{fig:cocktail-metrics}.  
In Fig. \ref{fig:cocktail-binary-matrices}, 
we plot the ground truth state
matrix against the average state matrix, $\boldeta^*$, 
averaged over runs and post-burn-in iterations.

The LT and Sticky-LT models outperform the others,
while the regular Sticky model exhibits only a small
advantage over the vanilla HDP-HMM.  Both converge on a non-negligible $\lambda$
value of about 1.6 (see Fig. \ref{fig:cocktail-metrics}), suggesting that the local transition structure
explains the data well.  The LT models also use more states than the non-LT
models, perhaps owing to the fact that the weaker transition prior of
the non-LT model is more likely to explain nearby similar observations
as a single persisting state, whereas the LT model places a higher
probability on transitioning to a new state with a similar latent vector.

\subsection{Synthetic Data Without Local Transitions}
\label{sec:synth-data-without}

\begin{figure}[tb]
\centering
  \centerline{\includegraphics[width = 0.72\columnwidth, height = 1.8cm]{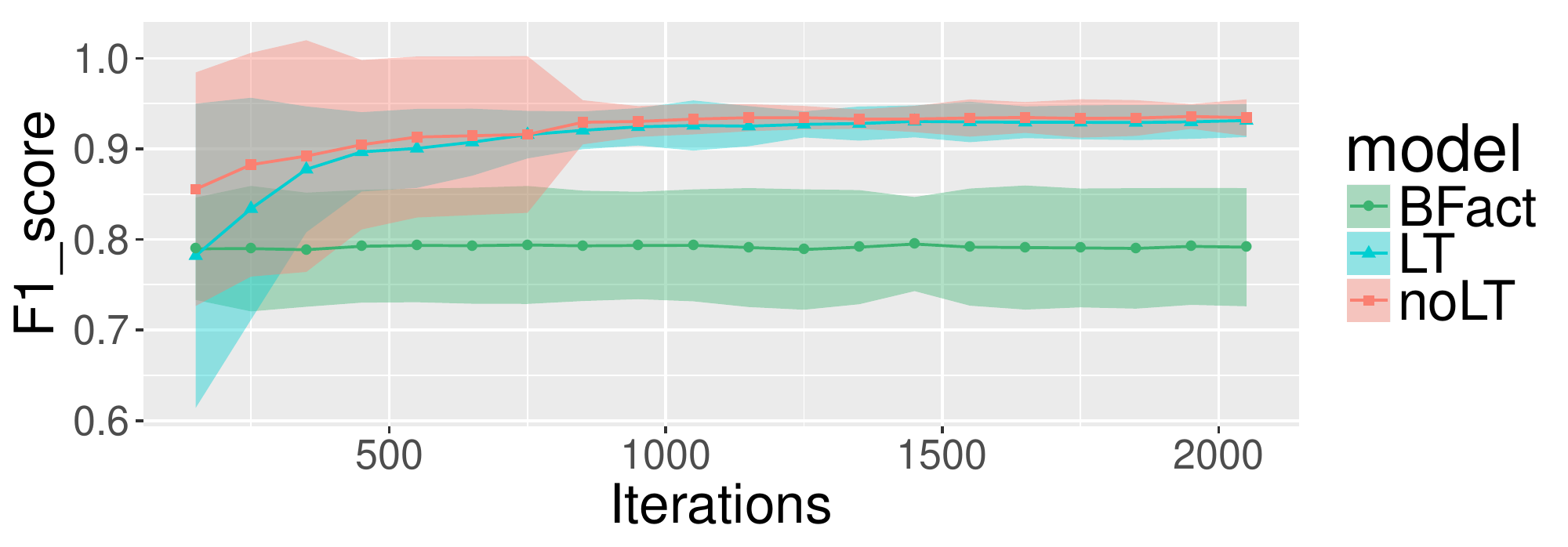}}
  \centerline{\includegraphics[width = 0.72\columnwidth, height = 1.8cm]{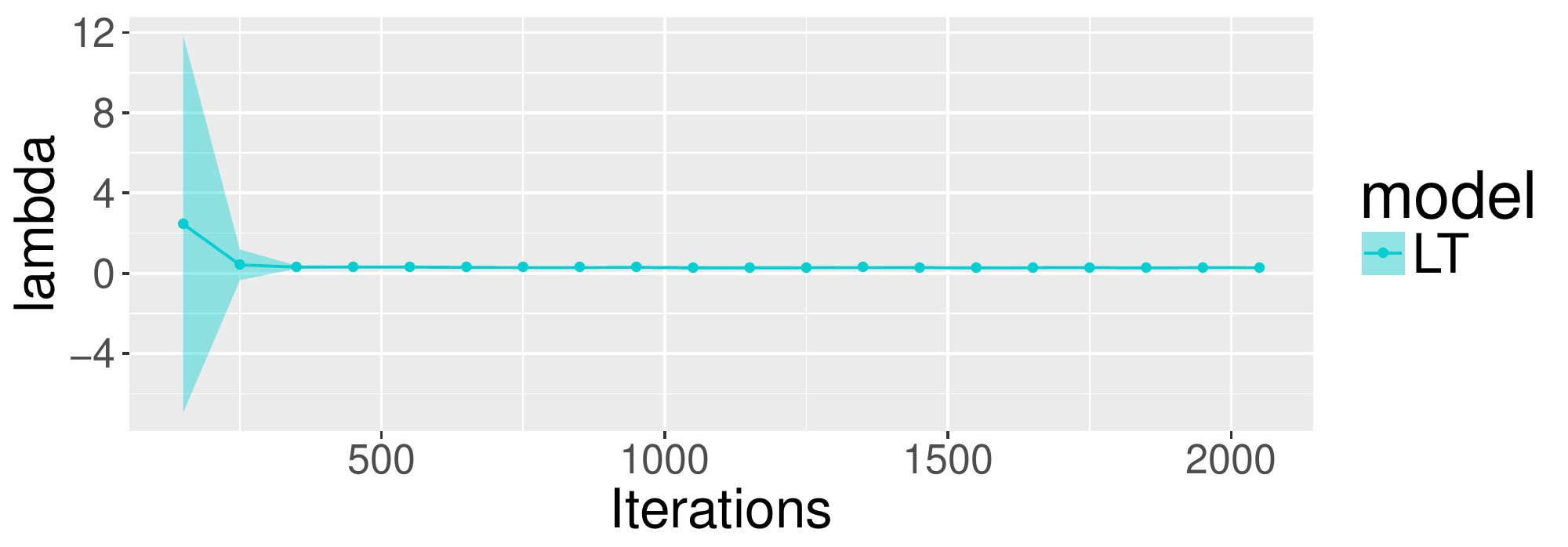}}
  \centerline{\includegraphics[width = 0.72\columnwidth, height = 1.8cm]{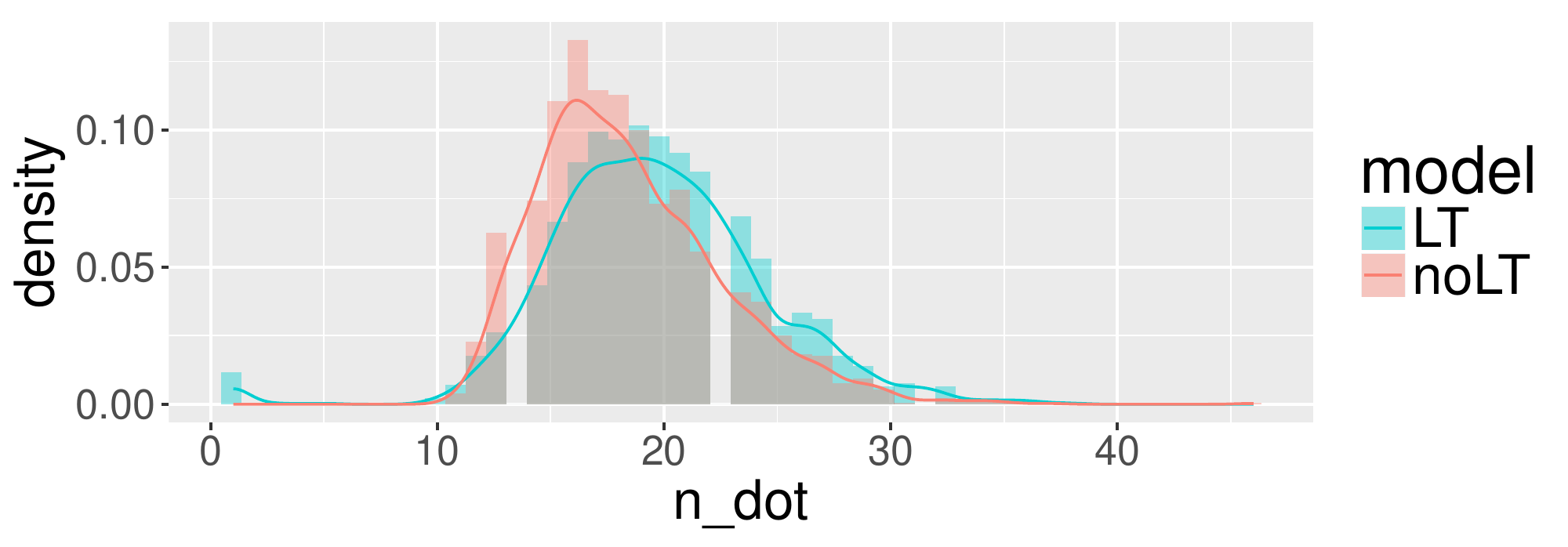}}
  \vspace{-0.2cm}
\caption{Top: F1 score for inferred relative to ground
  truth binary speaker matrices on synthetic data generated from the
  vanilla HDP-HMM model.  Middle: Learned similarity parameter, $\lambda$, for the LT
  model by Gibbs iteration, averaged over 5 runs.  Bottom: Number of
  states used, $n_{\cdot}$, by each model in the training set. 
  Error bands are
  99\% confidence interval of the mean per iteration.  The first 100
  iterations are omitted. \label{fig:synthetic-metrics}}
\end{figure}

We generated data directly from the ordinary HDP-HMM used in the
cocktail experiment as a sanity check, to examine
the performance of the LT model in the absence of a similarity bias.
The results are in
Fig. \ref{fig:synthetic-metrics}.  When the $\lambda$ parameter is
large, the LT model has worse
performance than the non-LT model on this data;
however, the $\lambda$ parameter settles near zero as the 
model learns that local transitions are not more
probable.  When $\lambda = 0$, the HDP-HMM-LT is an ordinary HDP-HMM.
The LT model does not make entirely the same inferences as the non-LT
model, however; in particular, the $\alpha$ concentration parameter is
larger.  To some extent, $\alpha$ and $\lambda$ trade off: sparsity of
the transition matrix can be achieved either by beginning with a
sparse rate matrix prior to rescaling ($\alpha$ small), or by
beginning with a less sparse rate matrix which becomes sparser through
rescaling (larger $\alpha$ and non-zero $\lambda$).

\subsection{Bach Chorales}

To test a version of the HDP-HMM-LT model in which the components of
the latent state governing similarity are unrelated to the emission
distributions, we used our model to do unsupervised ``grammar''
learning from a corpus of Bach chorales.  The data was a corpus of 217 four-voice
major key chorales by J.S. Bach from
music21\footnote{\scriptsize \url{http://web.mit.edu/music21}}, 
200 of which were randomly selected
as a training set, with the other 17 used as a test set to evaluate
surprisal (marginal log likelihood per observation) by the trained
models.  All chorales were transposed to C-major, and each
distinct four-voice chord (with voices ordered) 
was encoded as a single integer.  In total
there were 3307 distinct chord types and 20401 chord tokens in the 217
chorales, with 3165 types and 18818 tokens in the 200 training
chorales, and 143 chord types that were unique to the test set.

\paragraph{Modifications to Model and Inference}
\label{sec:model-inference}

Since the chords were encoded as integers, the emission distribution
for each state is $\Cat{\btheta_j}$.  We use a
symmetric Dirichlet prior for each $\btheta_j$, resulting in conjugate 
updates to $\btheta$ conditioned on the latent state sequence, $\bz$.

In this experiment, the locations, $\bell_j$, are
independent of the $\btheta_j$, with
$\Norm{0}{\mathbf{I}}$ priors.  We use a Gaussian
similarity function, $\phi_{jj} :=
\exp\{-\lambda d_2(\bell_j, \bell_{j'})^2\}$ where $d_2$ is Euclidean
distance.  Since the latent
states are continuous, we use a Hamiltonian Monte Carlo (HMC) update
\cite{duane1987hybrid, neal2011mcmc} to update the $\bell_j$
simultaneously, conditioned on $\bz$ and $\bpi$ (see Appendix \ref{sec:hmc}
for details).  

\paragraph{Results}
\label{sec:results}

\begin{figure}[tb]
\centering
  \centerline{\includegraphics[width = 0.75\columnwidth, height = 1.8cm]{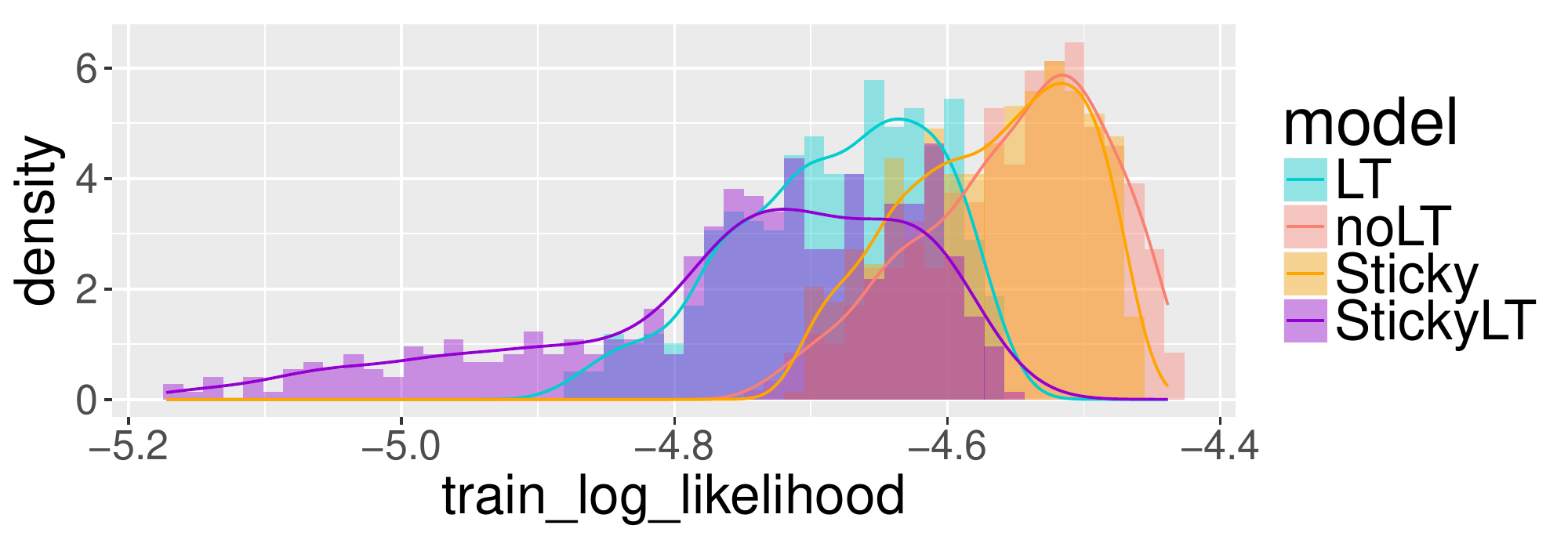}}
  \centerline{\includegraphics[width = 0.75\columnwidth, height = 1.8cm]{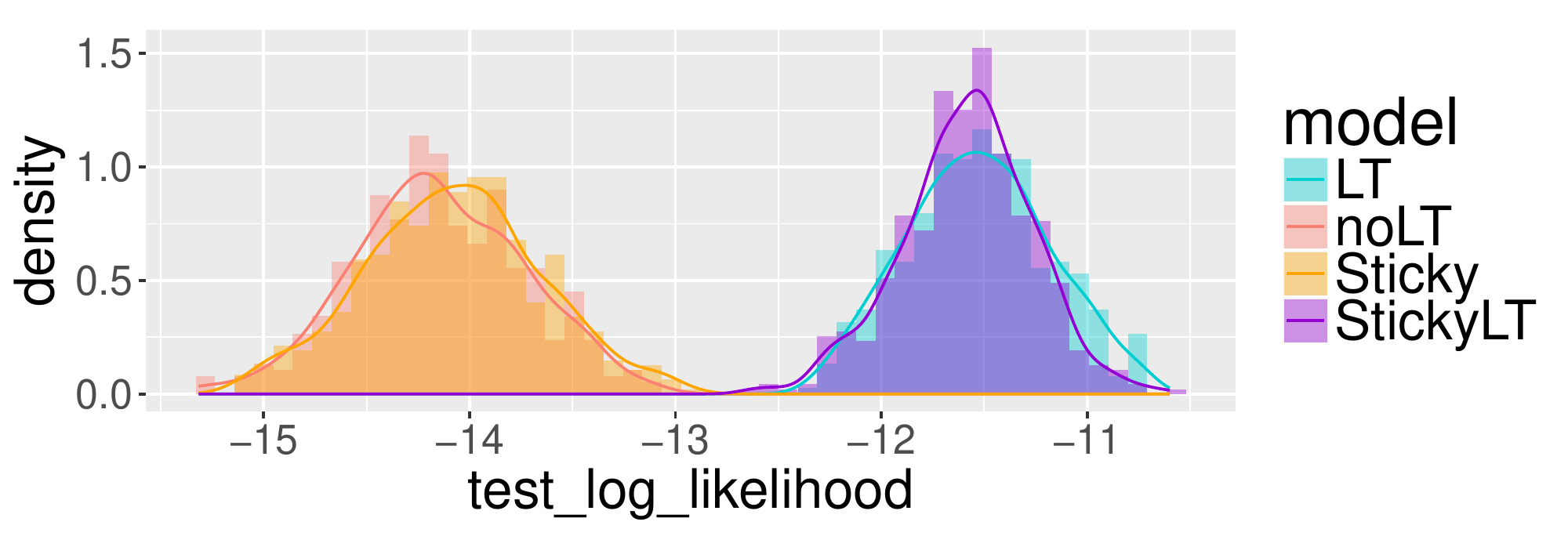}}
  \vspace{-0.2cm}
\caption{Training set and test set log marginal likelihoods for Bach
  chorale data on the four HDP-based models: HDP-HMM-LT, HDP-HMM,
  Sticky HMM, and Sticky HDP-HMM-LT. \label{fig:bach-metric}
}
\end{figure}

We ran 5 Gibbs chains for 10,000 iterations 
each using the HDP-HMM-LT, Sticky-HDP-HMM-LT, HDP-HMM and
Sticky-HDP-HMM models on the 200
training chorales, which were modeled as conditionally independent
of one another.  We evaluated the
marginal log likelihood on the 17 test chorales (integrating out $\bz$) 
at every 50th iteration.  The training and test log
likelihoods are in Fig. \ref{fig:bach-metric}.  Although the LT
model does not achieve as close a fit to the training data, its
generalization performance is better, suggesting that the vanilla HDP-HMM
is overfitting.  This is perhaps counterintuitive, since the LT model
is more flexible, and might be expected to be more prone to
overfitting.  However, the similarity bias induces greater information 
sharing across parameters, as in a hierarchical model: instead of each
entry of the transition matrix being informed mainly by transitions 
directly involving the corresponding states, it is informed to 
some extent by {\em all} transitions, as they all inform the similarity structure.

\section{Discussion}
\label{sec:discussion}

We have defined a new probabilistic model, 
the Hierarchical Dirichlet Process Hidden Markov Model
with Local Transitions (HDP-HMM-LT), which generalizes the HDP-HMM by
allowing state space geometry to be represented
via a similarity kernel, making transitions between ``nearby''
pairs of states (``local'' transitions), more likely {\em a priori}.  
By introducing an augmented data representation, which we call the
Markov Jump Process with Failed Transitions (MJP-FT), we obtain
a Gibbs sampling algorithm that simplifies inference in both the LT and
ordinary HDP-HMM.  When multiple latent chains are interdependent, as
in speaker diarization, the HDP-HMM-LT model
combines the HDP-HMM's capacity to discover a small set of joint states
with the Factorial HMM's ability to encode the property that most transitions involve
a small number of chains.  The HDP-HMM-LT outperforms both, as well as
outperforming the Sticky-HDP-HMM, on
a speaker diarization task in which speakers form conversational
groups.  Despite the addition of the similarity
kernel, the HDP-HMM-LT is able to suppress its local
transition prior when the data does not support it, achieving
identical performance to the HDP-HMM on data generated directly from
the latter.

The local transition property is particularly clear when transitions occur at
different times for different latent features, as with binary
vector-valued states in the cocktail party setting, 
but the model can be used with any state space 
equipped with a suitable similarity kernel.  Similarities need not be defined in
terms of emission parameters; state ``locations''
can be represented and inferred separately, which we demonstrate
using Bach chorale data.  There, the LT model achieves
better predictive performance on a held-out test set, while the
ordinary HDP-HMM overfits the training set: the LT
property here acts to encourage a concise harmonic representation
where chord contexts are arranged in bidirectional functional
relationships. 

We focused on fixed-dimension binary 
vectors for the cocktail party and synthetic
data experiments, but it would be straightforward 
to add the LT property to a model with nonparametric
latent states, such as the iFHMM \cite{gael2009infinite} and the infinite 
factorial dynamic model \cite{valera2015infinite}, both of which use 
the Indian Buffet Process (IBP) \cite{ghahramani2005infinite} as a state 
prior.  The similarity function used here could be employed without
changes: since only finitely many coordinates are non-zero in the IBP,
the distance between any two states is finite.

\subsubsection*{Acknowledgments}

This work was funded in part by DARPA grant W911NF-14-1-0395 under the Big Mechanism Program and DARPA grant W911NF-16-1-0567 under the Communicating with Computers Program.

\bibliographystyle{icml2017}
\bibliography{ms}
\appendix
\onecolumn
\section*{APPENDICES}
Appendix \ref{sec:mjp-ft} concerns the derivation of the augmented data
representation referred to as the ``Markov Jump Process with Failed
Transitions'' (MJP-FT).  Appendix \ref{sec:hdp-inference} fills in
details for the Gibbs sampling steps to sample the 
rescaled HDP used by the HDP-HMM-LT.  Appendix \ref{sec:cocktail-inference}
gives a derivation for the updates to the binary state vectors, $\btheta$, in the
version of the HDP-HMM-LT used in the cocktail party experiment.
Finally, appendix \ref{sec:hmc} gives the details for the Hamiltonian
Monte Carlo update for $\bell$ in the version of the model used in the
Bach chorale experiment.

\section{Details of the Markov Jump Process with Failed Transitions
  Representation}
\label{sec:mjp-ft}


We can gain stronger intuition, as well as simplify posterior
inference, by re-casting the HDP-HMM-LT
as a continuous time Markov Jump Process where some of the attempts to jump
from one state to another fail, and where the failure probability
increases as a function of the ``distance'' between the states.

Let $\phi$ be defined as in the last section, and let 
$\bbeta$, $\btheta$ and $\bpi$ be defined as in the Normalized Gamma
Process representation of the ordinary HDP-HMM.  That is,
\begin{align}
  \label{eq:beta} \bbeta &\sim \mathrm{GEM}(\gamma) \\
  \theta_j &\stackrel{i.i.d}{\sim} H \\
  \pi_{jj'} \given \bbeta, \btheta &\sim \Gamm{\alpha \beta_{j'}}{1}
\end{align}
Now suppose that when the process is in state $j$, jumps to state
$j'$ are made at rate $\pi_{jj'}$.  This defines a continuous-time
Markov Process where the off-diagonal elements of the transition rate
matrix are the off diagonal elements of $\bpi$.  In addition,
self-jumps are allowed, and occur with rate $\pi_{jj}$.   If we only
observe the jumps and not the durations between jumps, this is an
ordinary Markov chain, whose transition matrix is obtained by
appropriately normalizing $\bpi$.  If we do not observe the jumps themselves, but
instead an observation is generated once per jump from a distribution that depends
on the state being jumped to, then we have an ordinary HMM.

We modify this process as follows.  
Suppose that each jump attempt from state $j$ to state $j'$ has a
chance of failing, which is an increasing function of the ``distance''
between the states.  In particular, let the success probability be
$\phi_{jj'}$ (recall that we assumed above that $0 \leq \phi_{jj'}
\leq 1$ for all $j,j'$).  Then, the rate of successful jumps from $j$
to $j'$ is $\pi_{jj'}\phi_{jj'}$, and the corresponding rate of unsuccessful jump
attempts is $\pi_{jj'}(1-\phi_{jj'})$.  To see this, denote by
$N_{jj'}$ the total number of jump attempts to $j'$ in a unit
interval of time spent in state $j$.  Since we are assuming the
process is Markovian, the total number of attempts is $\Pois{\pi_{jj'}}$
distributed.  Conditioned on $N_{jj'}$, $n_{jj'}$ will be successful, where
\begin{equation}
  \label{eq:51}
  n_{jj'} \given N_{jj'} \sim \Binom{N_{jj'}}{\phi_{jj'}}
\end{equation}
It is easy to show (and well known) that the marginal distribution of
$n_{jj'}$ is $\Pois{\pi_{jj'}\phi_{jj'}}$, and the marginal
distribution of $\tilde{q}_{jj'} := N_{jj'} - n_{jj'}$ is
$\Pois{\pi_{jj'}(1-\phi_{jj'})}$.  The rate of successful jumps
from state $j$ overall is then $T_j := \sum_{j'} \pi_{jj'} \phi_{jj'}$.

Let $t$ index jumps, so that $z_t$ indicates the $t$th state visited
by the process (couting self-jumps as a new time step).  Given
that the process is in state $j$ at discretized time $t-1$ (that is,
$z_{t-1} = j$), it is a standard property of Markov Processes that 
the probability that the first successful jump is to state $j'$ (that is, $z_{t} = j'$) 
is proportional to the rate of successful attempts to 
$j'$, which is $\pi_{jj'}\phi_{jj'}$.  

Let $\tilde{u}_{t}$ indicate the time elapsed between the $t$th and 
and $t-1$th successful jump (where we assume that the first
observation occurs when the first successful jump from a distinguished initial
state is made).  We have
\begin{equation}
  \label{eq:52}
  \tilde{u}_t \given z_{t-1} \sim \Exp{T_{z_{t-1}}}
\end{equation}
where $\tilde{u}_t$ is independent of $z_{t}$.

During this period, there will be $\tilde{q}_{j't}$ unsuccessful attempts to
jump to state $j'$, where
\begin{equation}
  \label{eq:53}
  \tilde{q}_{j't} \given z_{t-1} \sim \Pois{\tilde{u}_t \pi_{z_{t-1}j'}(1-\phi_{z_{t-1}j'})}
\end{equation}

Define the following additional variables
\begin{align}
  \label{eq:56}
    \mathcal{T}_j &= \{t \given z_{t-1} = j\} \\
    q_{jj'} &= \sum_{t \in \mathcal{T}_j}
    \tilde{q}_{j't} \\
    u_j &= \sum_{t \in \mathcal{T}_j} \tilde{u}_t 
\end{align}
and let $\bQ = (q_{jj'})_{j,j' \geq 1}$ be the matrix of unsuccessful
jump attempt counts, and $\bu = (u_j)_{j \geq 1}$ be the vector of
the total times spent in each state.

Since each of the $\tilde{u}_t$ with $t \in \mathcal{T}_j$ are
i.i.d. $\Exp{T_j}$, we get the marginal distribution
\begin{equation}
u_j \given \bz, \bpi, \bphi \stackrel{ind}{\sim} \Gamm{n_{j\cdot}}{T_j}
\end{equation}
by the standard property that sums of i.i.d. Exponential distributions
has a Gamma distribution with shape equal to the number of variates in
the sum, and rate equal to the rate of the individual exponentials.  
Moreover, since the $\tilde{q}_{j't}$ with $t \in \mathcal{T}_j$ 
are Poisson distributed, the total number of failed
attempts in the total duration $u_j$ is
\begin{equation}
  \label{eq:60}
  q_{jj'} \stackrel{ind}{\sim} \Pois{u_j\pi_{jj'}(1-\phi_{jj'})}.
\end{equation}

Thus if we marginalize out the individual $\tilde{u}_t$ and
$\tilde{q}_{j't}$, we have a joint distribution
over $\bz$, $\bu$, and $\bQ$, conditioned on the transition rate
matrix $\bpi$ and the success probability matrix $\bphi$, which is
\begin{align}
  \label{eq:54}
  p(\bz, \bu, \bQ \given \bpi, \bphi) &= \left(\prod_{t=1}^T p(z_{t} \given
    z_{t-1})\right) \prod_{j} p(u_j \given \bz, \bpi, \bphi)
  \prod_{j'} p(q_{jj'} \given u_j \pi_{jj'}, \phi_{jj'}) \\
  &= \left(\prod_{t} \frac{\pi_{z_{t-1}z_t}\phi_{z_{t-1}z_t}}{T_{z_{t-1}}}\right) \prod_{j}
  \frac{T_j^{n_{j\cdot}}}{\Gamma(n_{j\cdot})} u_j^{n_{j\cdot} - 1}
  e^{-T_j u_j} \\ &\qquad\qquad\times
  \prod_{j'} e^{-u_j\pi_{jj'}(1-\phi_{jj'})} u_j^{q_{jj'}}
  \pi_{jj'}^{q_{jj'}} (1-\phi_{jj'})^{q_{jj'}} (q_{jj'}!)^{-1} \\
  &= \prod_{j} \Gamma(n_{j\cdot})^{-1} u_j^{n_{j\cdot} + q_{j\cdot}-1}
  \\ &\qquad\qquad \times \prod_{j'}
  \pi_{jj'}^{n_{jj'} + q_{jj'}} \phi_{jj'}^{n_{jj'}}
  (1-\phi_{jj'})^{q_{jj'}} e^{-\pi_{jj'}\phi_{jj'}u_j}
  e^{-\pi_{jj'}(1-\phi_{jj'})u_j} (q_{jj'}!)^{-1} \\
  &\label{eq:joint-likelihood} = \prod_{j} \Gamma(n_{j\cdot})^{-1} u_j^{n_{j\cdot} + q_{j\cdot}-1} \prod_{j'}
  \pi_{jj'}^{n_{jj'} + q_{jj'}} \phi_{jj'}^{n_{jj'}}
  (1-\phi_{jj'})^{q_{jj'}} e^{-\pi_{jj'}u_j} (q_{jj'}!)^{-1}
\end{align}
Setting aside terms that do not depend on $\bpi$, we get the
conditional likelihood function used in sampling $\bpi$:
\begin{align}
  p(\bz, \bu, \bQ \given \bpi, \bphi) &\propto
  \prod_{j}\prod_{j'}\pi_{jj'}^{n_{jj'}+q_{jj'}} e^{-\pi_{jj'}u_j}
\end{align}
which, combined with the independent Gamma priors on $\pi$ yields
conditionally independent Gamma posteriors:
\begin{align}
  \pi_{jj'} \given \bz, \bu, \bQ, \bbeta, \alpha \stackrel{\text{ind.}}{\sim}
  \Gamm{\alpha\bbeta_{j'} + n_{jj'} + q_{jj'}}{1 + u_j}
\end{align}

\section{Inference details for hyperparameters of the rescaled HDP}
\label{sec:hdp-inference}

\subsection{Sampling $\bpi$, $\bbeta$, $\alpha$ and $\gamma$}
\label{sec:sampling-pi}

The joint conditional over $\gamma$, $\alpha$, $\bbeta$ and $\bpi$ given
the augmented data $\mathcal{D} = (\bz, \bu, \bQ, \bM, \br, w)$
factors as
\begin{equation}
  \label{eq:46} p(\gamma, \alpha, \beta, \pi \given \mathcal{D}) = p(\gamma \given \mathcal{D}) p(\alpha \given \mathcal{D}) p(\beta \given
\gamma, \mathcal{D}) p(\pi \given \alpha, \beta, \mathcal{D})
\end{equation} We will derive these four factors in reverse order.

\paragraph{Sampling $\bpi$}

The entries in $\bpi$ are conditionally independent given $\alpha$ and
$\beta$, so we have the prior
\begin{equation}
  \label{eq:47} p(\bpi \given \bbeta, \alpha) = \prod_{j} \prod_{j'}
\Gamma(\alpha\bbeta_{j'})^{-1} \pi_{jj'}^{\alpha\bbeta_{j'} - 1} \exp(-\pi_{jj'}),
\end{equation} 
and the likelihood given $\{\bz, \bu, \bQ\}$
given by \eqref{eq:joint-likelihood}.  Combining these, we have
\begin{align}
  \label{eq:61} p(\bpi, \bz, \bu, \bQ \given \beta, \alpha, \bphi) &=
\prod_{j} u_j^{n_{j\cdot} + q_{j\cdot} - 1}\prod_{j'}
\Gamma(\alpha\beta_{j'})^{-1} \pi_{jj'}^{\alpha\beta_{j'} + n_{jj'} +
q_{jj'} - 1} \\&\qquad \times e^{-(1 + u_j) \pi_{jj'}}
\phi_{jj'}^{n_{jj'}} (1-\phi_{jj'})^{q_{jj'}} (q_{jj'}!)^{-1}
\end{align} 
Conditioning on everything except $\bpi$, we get
\begin{align}
  \label{eq:24} p(\bpi \given \bQ, \bu, \bz, \bbeta, \alpha) &\propto
\prod_j \prod_{j'} \pi_{jj'}^{\alpha\beta_{j'} + n_{jj'} + q_{jj'} -
1} \exp(-(1 + u_j)\pi_{jj'})
\end{align} 
and thus we see that the $\pi_{jj'}$ are conditionally
independent given $u$, $z$ and $Q$, and distributed according to
\begin{align}
  \label{eq:25} 
\pi_{jj'} \given n_{jj'}, q_{jj'}, \beta_{j'}, \alpha
\stackrel{ind}{\sim} \Gamm{\alpha\beta_{j'} + n_{jj'} + q_{jj'}}{1 +
u_j}
\end{align}

\paragraph{Sampling $\bbeta$}
\label{sec:sampling-bbeta}

Consider the conditional distribution of $\beta$ having integrated out
$\bpi$.  The prior density of $\bbeta$ is
\begin{equation}
  \label{eq:62} p(\bbeta \given \gamma) =
\frac{\Gamma(\gamma)}{\Gamma(\frac{\gamma}{J})^J} \prod_{j}
\beta_j^{\frac{\gamma}{J} - 1}
\end{equation} 
After integrating out $\pi$ in \eqref{eq:61}, we have
\begin{align} p(\bz, \bu, \bQ \given \bbeta, \alpha, \gamma, \bphi) &=
\prod_{j=1}^J u_{j} ^{-1} \prod_{j'=1}^J u^{n_{jj'} + q_{jj'} - 1}(1 +
u_j)^{-(\alpha\beta_{j'} + n_{jj'} + q_{jj'})} \\ &\qquad \qquad
\times \frac{\Gamma(\alpha\beta_{j'} + n_{jj'} +
q_{jj'})}{\Gamma(\alpha\beta_{j'})}
\phi_{jj'}^{n_{jj'}}(1-\phi_{jj'})^{q_{jj'}} (q_{jj'}!)^{-1} \\ &=
\prod_{j=1}^J \Gamma(n_{j\cdot})^{-1} u_j^{-1}(1+u_j)^{-\alpha}
\left(\frac{u_j}{1+u_j}\right)^{n_{j\cdot} + q_{j\cdot}} \\ &\qquad
\qquad \times \prod_{j' = 1}^J \frac{\Gamma(\alpha\beta_{j'} + n_{jj'}
+ q_{jj'})}{\Gamma(\alpha\beta_{j'})}
\phi_{jj'}^{n_{jj'}}(1-\phi_{jj'})^{q_{jj'}} (q_{jj'}!)^{-1}
\end{align} 
where we have used the fact that the $\beta_j$ sum to 1. Therefore
\begin{align} 
p(\bbeta \given \bz, \bu, \bQ, \alpha, \gamma) &\propto
\prod_{j=1}^J \beta_j^{\frac{\gamma}{J} - 1} \prod_{j'=1}^J
\frac{\Gamma(\alpha\beta_{j'} + n_{jj'} +
q_{jj'})}{\Gamma(\alpha\beta_{j'})}.
\end{align}

Following \cite{teh2006hierarchical}, we can write the ratios of
Gamma functions as polynomials in $\beta_j$, as
\begin{equation}
  \label{eq:31} p(\bbeta \given \bz, \bu, \bQ, \alpha, \gamma)
\propto \prod_{j=1}^J \beta_j^{\frac{\gamma}{J} - 1} \prod_{j'=1}^{J}
\sum_{m_{jj'} = 1}^{n_{jj'}} s(n_{jj'} + q_{jj'}, m_{jj'}) (\alpha
\beta_{j'})^{m_{jj'}}
\end{equation} 
where $s(m,n)$ is an unsigned Stirling number of the
first kind, which is used to represent the number of permutations of
$n$ elements such that there are $m$ distinct cycles.

This admits an augmented data representation, where we introduce a
random matrix $\bM = (m_{jj'})_{1 \leq j,j' \leq J}$, whose entries are
conditionally independent given $\bbeta$, $\bQ$ and $\bz$, with
\begin{equation}
  \label{eq:32} 
p(m_{jj'} = m \given \beta_{j'}, \alpha, n_{jj'},
q_{jj'}) = \frac{s(n_{jj'} + q_{jj'}, m) \alpha^{m}
\beta_{j'}^{m}}{\sum_{m'=0}^{n_{jj'} + q_{jj'}} s(n_{jj'} + q_{jj'},
m') \alpha^{m'} \beta_{j'}^{m'}}
\end{equation} 
for integer $m$ ranging between $0$ and $n_{jj'} +
q_{jj'}$.  Note that $s(n,0) = 0$ if $n > 0$, $s(0,0) = 1$, $s(0,m) =
0$ if $m > 0$, and we have the recurrence relation $s(n+1,m) = n
s(n,m) + s(n,m-1)$, and so we could compute each of these coefficients
explicitly; however, it is typically simpler and more computationally
efficient to sample from this distribution by simulating the number of
occupied tables in a Chinese Restaurant Process with $n$ customers, 
than it is to enumerate its probabilities.

For each $m_{jj'}$ we simply draw $n_{jj'}$ assignments of customers
to tables according to the Chinese Restaurant Process and set
$m_{jj'}$ to be the number of distinct tables realized; that is,
assign the first customer to a table, setting $m_{jj'}$ to 1, and
then, after $n$ customers are assigned, assign the $n+1$th customer to
a new table with probability $\alpha\beta_{j'} / (n +
\alpha\beta_{j'})$, in which case we increment $m_{jj'}$, and to an
existing table with probability $n / (n + \alpha)$, in which case we
do not increment $m_{jj'}$.



Then, we have joint distribution
\begin{equation}
  \label{eq:33} p(\bbeta, \bM \given \bz, \bu, \bQ, \alpha, \gamma)
\propto \prod_{j=1}^J \beta_j^{\frac{\gamma}{J} - 1} \prod_{j'=1}^{J}
s(n_{jj'} + q_{jj'}, m_{jj'}) \alpha^{m_{jj'}} \beta_{j'}^{m_{jj'}}
\end{equation} 
which yields \eqref{eq:31} when marginalized over $\bM$.
Again discarding constants in $\bbeta$ and regrouping yields
\begin{equation}
  \label{eq:34} p(\beta \given M, z, u, \theta, \alpha, \gamma)
\propto \prod_{{j'}=1}^J \beta_{j'}^{\frac{\gamma}{J} + m_{\cdot
{j'}}- 1}
\end{equation} which is Dirichlet:
\begin{equation}
  \label{eq:38} \beta \given M, \gamma \sim
\mathrm{Dirichlet}(\frac{\gamma}{J} + m_{\cdot 1}, \dots,
\frac{\gamma}{J} + m_{\cdot J})
\end{equation}

\paragraph{Sampling $\alpha$ and $\gamma$}
\label{sec:sampling-alpha} Assume that $\alpha$ and $\gamma$ have
Gamma priors, parameterized by shape, $a$ and rate, $b$:
\begin{align}
  \label{eq:42} p(\alpha) &=
\frac{b_{\alpha}^{a_{\alpha}}}{\Gamma(a_{\alpha})} \alpha^{a_{\alpha}
- 1} \exp(-b_{\alpha}\alpha) \\ p(\gamma) &=
\frac{b_{\gamma}^{a_\gamma}}{\Gamma(a_{\gamma})} \gamma^{a_{\gamma -
1}} \exp(-b_{\gamma}\gamma)
\end{align}

Having integrated out $\bpi$, we have
\begin{align} 
p(\bbeta, \bz, \bu, \bQ, \bM \given \alpha, \gamma, \bphi) &=
\frac{\Gamma(\gamma)}{\Gamma(\frac{\gamma}{J})^J}
\alpha^{m_{\cdot\cdot}} \prod_{j=1}^J \beta_j^{\frac{\gamma}{J} +
m_{\cdot j} - 1}\Gamma(n_{j\cdot})^{-1} u_j^{-1}(1+u_j)^{-\alpha}
\left(\frac{u_j}{1+u_j}\right)^{n_{j\cdot} + q_{j\cdot}} \\ &\qquad
\qquad \times \prod_{j' = 1}^J s(n_{jj'} + q_{jj'}, m_{jj'})
\phi_{jj'}^{n_{jj'}}(1-\phi_{jj'})^{q_{jj'}} (q_{jj'}!)^{-1}
\end{align} 
We can also integrate out $\bbeta$, to yield
\begin{align} 
p(\bz, \bu, \bQ, \bM \given \alpha, \gamma, \bphi) &=
\alpha^{m_{\cdot\cdot}} e^{-\sum_{j''} \log(1+u_{j''}) \alpha}
\frac{\Gamma(\gamma)}{\Gamma(\gamma + m_{\cdot\cdot})} \\ &\qquad
\qquad \times \prod_j \frac{\Gamma(\frac{\gamma}{J} + m_{\cdot
j})}{\Gamma(\frac{\gamma}{J}) \Gamma(n_{j\cdot})} u_j^{-1}
\left(\frac{u_j}{1+u_j}\right)^{n_{j\cdot} + q_{j\cdot}} \\ &\qquad
\qquad \times \prod_{j' = 1}^J s(n_{jj'} + q_{jj'}, m_{jj'})
\phi_{jj'}^{n_{jj'}}(1-\phi_{jj'})^{q_{jj'}} (q_{jj'}!)^{-1}
\end{align} 
demonstrating that $\alpha$ and $\gamma$ are independent
given $\bphi$ and the augmented data, with
\begin{equation}
  \label{eq:43} p(\alpha \given \bz, \bu, \bQ, \bM) \propto
\alpha^{a_{\alpha} + m_{\cdot\cdot}}\exp(-(b_\alpha +
\sum_{j}\log(1+u_j))\alpha)
\end{equation} and
\begin{align}
  \label{eq:8} p(\gamma \given \bz, \bu, \bQ, \bM) &\propto
\gamma^{a_{\gamma - 1}} \exp(-b_{\gamma}\gamma)
\frac{\Gamma(\gamma)\prod_{j=1}^J \Gamma(\frac{\gamma}{J} + m_{\cdot
j})}{\Gamma(\frac{\gamma}{J})^J\Gamma(\gamma + m_{\cdot\cdot})}
\end{align} 
So we see that
\begin{equation}
  \label{eq:44} \alpha \given \bz, \bu, \bQ, \bM \sim \Gamm{a_{\alpha}
+ m_{\cdot\cdot}}{b_\alpha + \sum_j\log(1+u_j)}
\end{equation} To sample $\gamma$, we introduce a new set of auxiliary
variables, $\br = (r_1, \dots, r_J)$ and $w$ with the following
distributions:
\begin{align}
  \label{eq:9} p(r_{j'} = r \given m_{\cdot {j'}}, \gamma) &=
\frac{\Gamma(\frac{\gamma}{J})}{\Gamma(\frac{\gamma}{J} + m_{\cdot
{j'}})} s(m_{\cdot {j'}}, r) \left(\frac{\gamma}{J}\right)^r \qquad r
= 1, \dots, m_{\cdot j} \\ p(w \given m_{\cdot\cdot} \gamma) &=
\frac{\Gamma(\gamma + m_{\cdot\cdot})}{\Gamma(\gamma)
\Gamma(m_{\cdot\cdot})} w^{\gamma - 1} (1-w)^{m_{\cdot\cdot} - 1}
\qquad w \in (0,1)
\end{align} so that
\begin{align}
  \label{eq:10} p(\gamma, r, w \given \bM) &\propto \gamma^{a_{\gamma -
1}} \exp(-b_{\gamma}\gamma) w^{\gamma - 1}(1-w)^{m_{\cdot\cdot} - 1}
\prod_{j'=1}^J s(m_{\cdot {j'}}, r_{j'})
\left(\frac{\gamma}{J}\right)^{r_{j'}}
\end{align} and
\begin{align}
  \label{eq:11} p(\gamma \given r, w) \propto \gamma^{a_\gamma +
r_{\cdot} - 1} \exp(-(b_{\gamma} - \log(w)) \gamma),
\end{align} which is to say
\begin{equation}
  \label{eq:18} \gamma \given \br, w, \bz, \bu, \bQ, \bM \sim
\Gamm{a_{\gamma} + r_{\cdot}}{b_{\gamma} - \log(w)}
\end{equation}

\section{Derivation of $\boldeta$ update in the Cocktail Party and
  Synthetic Data Experiments}
\label{sec:cocktail-inference}

In principle, $\boldeta$ can have any distribution over binary vectors, but we
will suppose for simplicity that it can be factored into $D$
independent coordinate-wise Bernoulli variates.  Let $\mu_d$ be the
Bernoulli parameter for the $d$th coordinate.

The similarity function $\phi_{jj'}$ is the Laplacian kernel:
\begin{align}
  \label{eq:39}
  \phi_{jj'} &= \Phi(\boldeta_j, \boldeta_{j'}) = \exp(-\lambda d_{jj'})
\end{align}
where $d_{jj'd} = \abs{\boldeta_{jd} - \boldeta_{j'd}}$ is
Hamming distance in the $d$th coordinate, $d_{jj'} :=
\sum_{d=1}^D d_{jj'}$ is the total Hamming
distance between $\boldeta_j$ and $\boldeta_{j'}$,
and $\lambda \geq 0$ (if $\lambda = 0$, the $\phi_{jj'}$
are identically 1, and so do not have any influence, reducing the
model to an ordinary HDP-HMM).

Let
\begin{align}
  \label{eq:68}
  \phi_{jj'-d} &= \exp(-\lambda(d_{jj'} - d_{jj'd}))
\end{align}
so that $\phi_{jj'} = \phi_{jj'-d} e^{-\lambda d_{jj'd}}$.

Since the matrix $\bphi$ is assumed to be symmetric, we have
\begin{align}
  \label{eq:70}
  \frac{p(\bz, \bQ \given \eta_{jd}  = 1, \boldeta\setminus\eta_{jd}
    )}{p(\bz, \bQ \given \eta_{jd}  = 0, \boldeta\setminus\eta_{jd} )}
  &\propto \prod_{j' \neq j}
  \frac{e^{-\lambda(n_{jj'} + n_{j'j})\abs{1 - \theta_{j'd}}}(1 -
    \phi_{jj'-d} e^{-\lambda\abs{1 - \theta_{j'd}}})^{q_{jj'} +
      q_{j'j}}}{e^{-\lambda(n_{jj'} + n_{j'j})\abs{\theta_{j'd}}}
    (1-\phi_{jj'-d}e^{-\lambda\abs{\theta_{j'd}}})^{q_{jj'} +
      q_{j'j}}} \\
  &= \label{eq:71} e^{-\lambda(c_{jd0} - c_{jd1})}
  \prod_{j' \neq j} \left(\frac{1 - \phi_{jj'-d}e^{-\lambda}}{1-\phi_{jj'-d}}\right)^{(-1)^{\theta_{j'd}}(q_{jj'} + q_{j'j})}
\end{align}
where $c_{jd0}$ and $c_{jd1}$ are the number of successful jumps to or
from state $j$, to or from states with a 0 or 1, respectively, in
position $d$.  That is,
\begin{equation}
  \label{eq:72}
  c_{jd0} = \sum_{\{j' \given \theta_{j'd} = 0\}} n_{jj'} + n_{j'j}\qquad c_{jd1} = \sum_{\{j' \given \theta_{j'd} = 1\}} n_{jj'} + n_{j'j}
\end{equation}

Therefore, we can Gibbs sample $\eta_{jd}$ from its conditional
posterior Bernoulli distribution given the rest of $\boldeta$, where
we compute the Bernoulli parameter via the log-odds
\begin{align}
  \label{eq:77}
  &\log\left(\frac{p(\eta_{jd} = 1 \given \bY, \bz, \bQ, \boldeta \setminus
    \eta_{jd})}{p(\eta_{jd} = 0 \given \bY, \bz, \bQ, \boldeta
    \setminus \eta_{jd})}\right) = \log\left(\frac{p(\eta_{jd} =
  1) p(\bz, \bQ \given \eta_{jd} = 1, \boldeta \setminus
  \eta_{jd}) p(\bY \given \bz, \eta_{jd} = 1, \boldeta \setminus \eta_{jd})}{p(\eta_{jd} = 0) p(\bz, \bQ \given \eta_{jd} = 0,
  \boldeta \setminus \eta_{jd}) p(\bY \given \bz, \eta_{jd} = 0,
  \boldeta \setminus \eta_{jd})}\right) \\ & \qquad = \log\left(\frac{\mu_d}{1 - \mu_d}\right)
  + (c_{jd1} - c_{jd0}) \lambda +
    \sum_{j' \neq j}
  (-1)^{\theta_{j'd}}(q_{jj'} + q_{j'j})\log\left(\frac{1 -
      \phi_{jj'}^{(-d)}e^{-\lambda}}{1-\phi_{jj'}^{(-d)}}\right) \\ &
  \qquad \qquad + \sum_{\{t \given z_t = j\}} \log\left(\frac{f(\by_t;
      \eta_{jd} = 1, \boldeta_j \setminus \eta_{jd})}{f(\by_t;
      \eta_{jd} = 0, \boldeta_j \setminus \eta_{jd})}\right)
\end{align}

Suppose also that the observed data $\bY$ consists of a $T \times K$
matrix, where the $t$th row $\by_t = (y_{t1}, \dots,
y_{tK})^{\mathsf{T}}$ is a $K$-dimensional feature vector associated
with time $t$, and let $\bW$ be a $D \times K$ weight matrix
with $k$th column $\bw_k$, such that 
\begin{equation}
  \label{eq:74}
  f(\by_t; \boldeta_j) = g(\by_t; \bW^{\mathsf{T}} \boldeta_j)
\end{equation}
for a suitable parametric function $g$.  We assume for simplicity
that $g$ factors as
\begin{equation}
  \label{eq:73}
  g(\by_t; \bW^{\mathsf{T}} \boldeta_j) = \prod_{k=1}^K g_k(y_{tk}; \bw_k \cdot \boldeta_j)
\end{equation}
Define $x_{tk} = \bw_k \cdot \theta_{z_{t}}$, and
$x_{tk}^{(-d)} = \bw_k^{-d} \cdot \theta_{z_{t}}^{-d}$, where
$\theta_{j}^{-d}$ and $\bw_{k}^{-d}$ are $\theta_{j}$ and $\bw_k$, respectively, with
the $d$th coordinate removed.  Then
\begin{equation}
  \label{eq:76}
  \log\left(\frac{f(\by_t; \eta_{jd} = 1, \boldeta_j \setminus
    \eta_{jd})}{f(\by_t; \eta_{jd} = 0, \boldeta_j \setminus \eta_{jd})}\right) =
  \sum_{k=1}^K \log\left(\frac {g_k(y_{tk};
    x_{tk}^{(-d)} + w_{dk})}{g_k(y_{tk};
    x_{tk}^{(-d)})}\right).
\end{equation}
If $g_k(y; x)$ is a Normal density with mean $x$ and unit variance, then
\begin{equation}
  \label{eq:91}
  \log\left(\frac {g_k(y_{tk};
    x_{tk}^{(-d)} + w_{dk})}{g_k(y_{tk};
    x_{tk}^{(-d)})}\right) = -w_{dk}(y_{tk} - x_{tk}^{(-d)} + \frac{1}{2}w_{dk})
\end{equation}

\section{Derivation of HMC update for $\bell$ in the Bach Chorale Experiment}
\label{sec:hmc}

We have a set of states with parameters $\bell_j$, $j = 1, \dots,
J$.  In the previous version of the model, $\bell_j$ was
a binary state vector on which both the similarities $\phi_{jj'}$
and the emission distribution $F_{j}$ depended.
Here, we define the latent locations $\bell_j = (\ell_{j1},
\ell_{jD})$ to be locations in $\mathbb{R}^D$, independent of
the emission distributions, so that during inference they are informed
solely by the transitions.

We set
\begin{equation*}
  \phi_{jj'}(\bell_j, \bell_{j'}) = \exp\left(-\frac{\lambda}{2} d_{jj'}^2\right)
\end{equation*}
where $d_{jj'}$ is the Euclidean distance between $\bell_j$ and
$\bell_{j'}$; that is,
\begin{equation*}
  d_{jj'}^2 = \sum_{d} (\ell_{jd} - \ell_{j'd})^2
\end{equation*}

Since now $\bell_j$ are continuous locations, we use Hamlitonian Monte
Carlo \cite{duane1987hybrid,
  neal2011mcmc} to sample them jointly.  HMC is a variation on Metropolis-Hastings
algorithm which is designed to more efficiently explore a
high-dimensional continuous distribution by adopting a proposal
distribution which incorporates an auxiliary ``momentum'' variable to
make it more likely that proposals will go in useful directions and
improve mixing compared to naive movement.  

To do Hamiltonian Monte Carlo to sample from the conditional posterior
of $\bell$ given $\bz$ and $\bQ$, we need to compute the gradient of the
log posterior, which is just the sum of the gradient of the log prior
and the gradient of the log likelihood.

Assume independent
and isotropic Gaussian priors on each $\bell_{j}$, so we have
\begin{equation*}
  p(\bell_j) \propto \exp\left(-\frac{h_\ell}{2} \sum_{d} \ell_{jd}^2 \right),
\end{equation*}
where $h_{\ell}$ is the prior precision which does not depend on $d$.

Then the log prior density, up to an additive constant $c$, is
\begin{equation*}
  \log p(\bell_j) = c - \frac{h_{\ell}}{2} \sum_{d} \ell_{jd}^2
\end{equation*}

The relevant log likelihood is
the log of the probability of the $\bz$ and $\bQ$ variables given the
$\phi_{jj'}$.  In particular, we have
\begin{equation*}
  L := p(\bz, \bQ \given \bphi) \propto \prod_{j} \prod_{j'} \phi_{jj'}^{n_{jj'}}(1 - \phi_{jj'})^{q_{jj'}}
\end{equation*}
so that
\begin{equation*}
  \log L = \sum_{j} \sum_{j'} \left( n_{jj'} \log(\phi_{jj'}) +
    q_{jj'} \log(1 - \phi_{jj'})\right)
\end{equation*}

The $j,d$ coordinate of the gradient of the log prior is simply
$-h_{\ell} \ell_{jd}$.

To get the $j,d$ coordinate of the gradient of the log likelihood, we
can apply the chain rule to terms as is convenient.  In particular,

\begin{equation*}
  \frac{\partial L}{\partial \ell_{jd}}=\sum_{j} \sum_{j'} n_{jj'}
  \frac{\partial \log(\phi_{jj'})}{\partial d_{jj'}^2}
  \frac{\partial d_{jj'}^2}{\partial \ell_{jd}}+ \sum_{j}
  \sum_{j'} q_{jj'} \frac{\partial \log(1 - \phi_{jj'})}{\partial (1 -
    \phi_{jj'})}\frac{\partial (1 - \phi_{jj'})}{\partial
    d_{jj'}^2} \frac{\partial d_{jj'}^2}{\partial \ell_{jd}}
\end{equation*}

We have the following components:
\begin{align*}
  \frac{\partial \log(\phi_{jj'})}{\partial d_{jj'}^2} &=
  -\frac{\lambda}{2} \\
  \frac{\partial d_{jj'}^2}{\partial \ell_{jd}} &= 2d_{jj'd} I(j \neq j')\\
  \frac{\partial \log(1 - \phi_{jj'})}{\partial (1 -
    \phi_{jj'})} &= \frac{1}{1 - \phi_{jj'}} \\
  \frac{\partial (1 - \phi_{jj'})}{\partial
    d_{jj'}^2} &= \frac{\lambda}{2} \phi_{jj'}
\end{align*}
which yields
\begin{align*}
  \frac{\partial L}{\partial \ell_{jd}} &= -\lambda \sum_{j}
  \sum_{j'} n_{jj'} d_{jj'd} \mathbb{I}(j \neq j') +
  \lambda \sum_{j}\sum_{j'} q_{jj'} d_{jj'd} \frac{\phi_{jj'}}{1 -
    \phi_{jj'}} \mathbb{I}(j \neq j) \\
  &= - \lambda \sum_{(j,j'): j \neq j'} d_{jj'd} \left(n_{jj'} - q_{jj'}
    \frac{\phi_{jj'}}{1 - \phi_{jj'}}\right)
\end{align*}

\end{document}